\RequirePackage[svgnames]{xcolor}

\documentclass[11pt,letterpaper]{mystyle}
\usepackage[all]{hypcap}
\usepackage[svgnames]{xcolor}
\usepackage[comma,authoryear,compress]{natbib}
\usepackage{hyperref}[citecolor=lightblue]

\hypersetup{
    colorlinks = true,
    citecolor = {YaleBlue},
}

\usepackage{algorithm}
\usepackage{algorithmicx}
\usepackage{algpseudocode}
\usepackage{microtype}
\usepackage{graphicx}
\expandafter\def\csname ver@subfig.sty\endcsname{}
\usepackage{booktabs} %
\usepackage{float}
\usepackage{bigstrut}

\usepackage{amsmath}
\usepackage{amssymb}
\usepackage{mathtools}
\usepackage{amsthm}
\usepackage{mathrsfs}
\usepackage{nicefrac}
\usepackage{dsfont}
\usepackage{subcaption}
\usepackage{graphicx,subfig}
\usepackage{cleveref}
\usepackage{bxcoloremoji}
\usepackage{listings}
\usepackage{tcolorbox}
\tcbuselibrary{listings}

\newtcblisting{promptbox}[2][]{
    listing only,
    listing options={
        basicstyle=\ttfamily\footnotesize,
        breaklines=true,
    },
    colback=gray!10,
    colframe=YaleBlue!60, 
    title=#2,
    fonttitle=\bfseries,
    #1
}
\usepackage{enumitem}
\setlist[itemize]{leftmargin=*}
\setlist[enumerate]{leftmargin=*}
\usepackage[normalem]{ulem}

\usepackage{float}
\usepackage{multirow}
\usepackage{afterpage}

\usepackage[utf8]{inputenc} %
\usepackage[T1]{fontenc}    %
\usepackage{hyperref}       %
\usepackage{url}            %
\usepackage{booktabs}       %
\usepackage{amsfonts}       %
\usepackage{nicefrac}       %
\usepackage{microtype}      %
\usepackage{graphicx}
\usepackage{subcaption} 
\usepackage{amssymb}
\usepackage{fdsymbol}
\usepackage{wrapfig}
\usepackage{lipsum}
\usepackage{stackengine}
\usepackage[font=small,labelfont=bf]{caption}
\usepackage{color}
\usepackage{adjustbox}

\usepackage{rotating}
\usepackage{makecell}
\usepackage{listings}  
\newcommand{\x}{\mathbf{x}}

\definecolor{blanchedalmond}{rgb}{1.0, 0.92, 0.8}
\definecolor{carmine}{rgb}{0.59, 0.0, 0.09}
\definecolor{lightblue}{rgb}{0.22,0.45,0.70}%

\renewcommand{\mathbf}{\boldsymbol}

\makeatletter
\def\Ddots{\mathinner{\mkern1mu\raise\p@
\vbox{\kern7\p@\hbox{.}}\mkern2mu
\raise4\p@\hbox{.}\mkern2mu\raise7\p@\hbox{.}\mkern1mu}}
\makeatother

\definecolor{amaranth}{rgb}{0.9, 0.17, 0.31}
\definecolor{antiquebrass}{rgb}{0.8, 0.58, 0.46}
\definecolor{antiquefuchsia}{rgb}{0.57, 0.36, 0.51}
\definecolor{chromeyellow}{rgb}{0.31, 0.47, 0.26}

\newcommand{\github}{\raisebox{-1.5pt}{\includegraphics[height=1.05em]{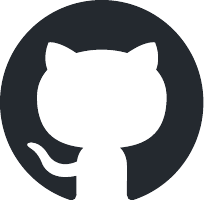}}}
\newcommand{\huggingface}{\raisebox{-1.5pt}{\includegraphics[height=1.05em]{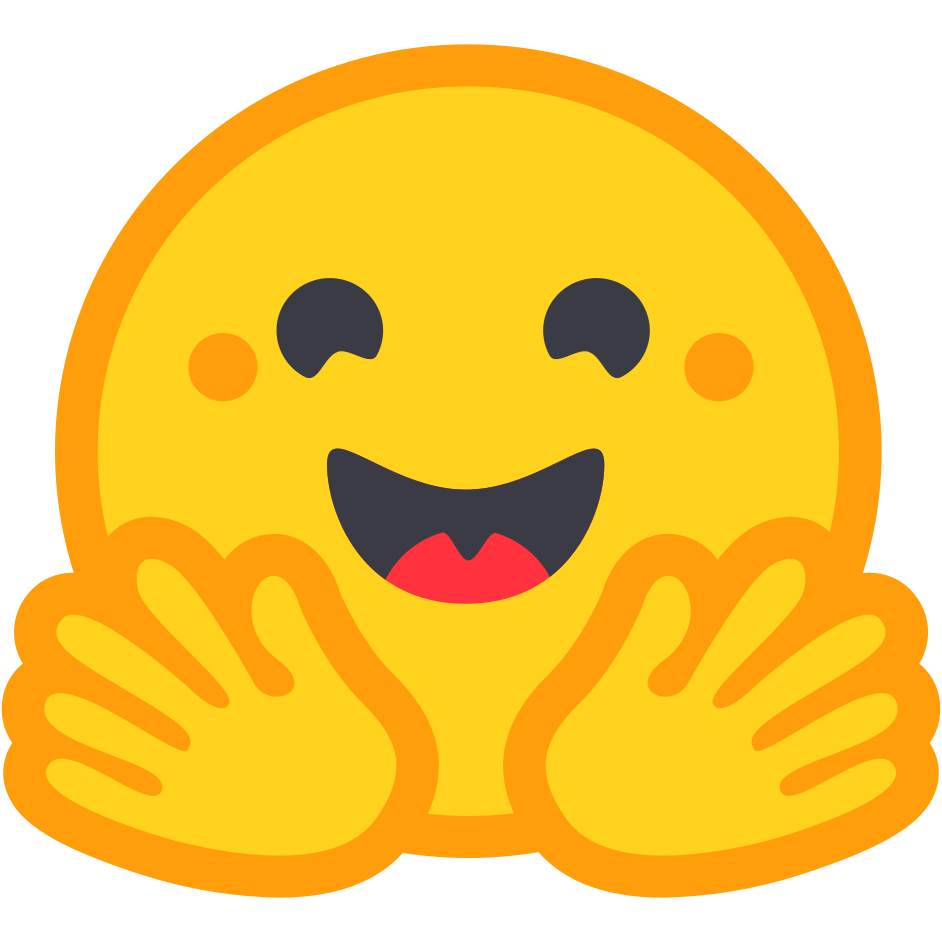}}}
\newcommand{\email}{\raisebox{-1.5pt}{\includegraphics[height=1.05em]{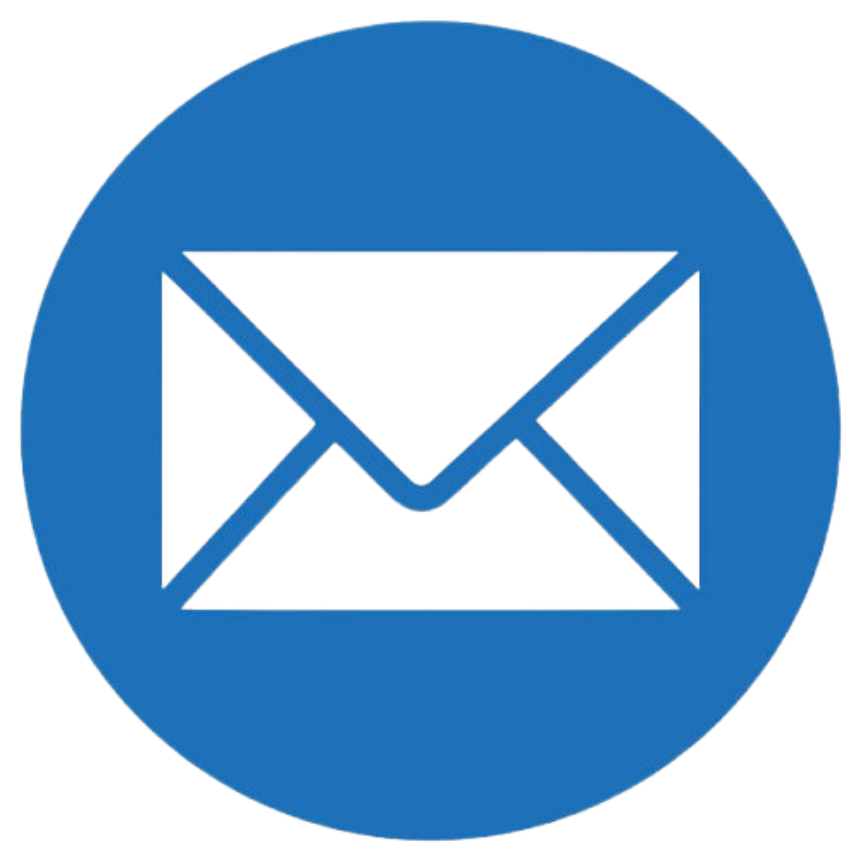}}}
\newcommand{\homepage}{\raisebox{-1.5pt}{\includegraphics[height=1.05em]{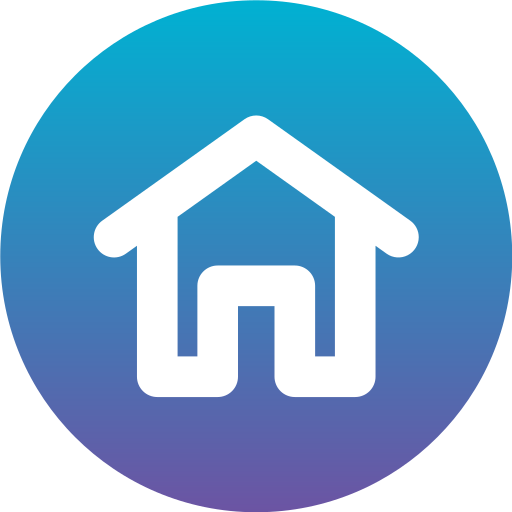}}}

\newtcolorbox{AIbox}[2][]{aibox,title=#2,#1}
\definecolor{lightblue}{rgb}{0.22,0.45,0.70}%
\definecolor{Gray}{gray}{0.95}
\definecolor{Cornsilk}{rgb}{1.0, 0.97, 0.86}

\usepackage{amsmath}

\usepackage{pdfrender}
\newcommand{\sota}[1]{%
  \text{\bfseries
    \pdfrender{StrokeColor=black,TextRenderingMode=2,LineWidth=0.3pt}{#1}}%
}

\title{\name: Benchmarking Reciprocal Cross-Modal Reasoning for Omnimodal Generation}
\runningtitle{\name: Benchmarking Reciprocal Cross-Modal Reasoning for Omnimodal Generation}

\author[$\triangle$]{Yongyuan Liang$^{*}$}
\author[$\blacktriangle$]{Wei Chow$^{*}$}
\author[$\lozenge$]{Feng Li}
\author[$\clubsuit$]{Ziqiao Ma}
\author[$\triangle$]{Xiyao Wang}
\author[$\bigstar$]{Jiageng Mao}
\author[$\triangle$]{\\Jiuhai Chen}
\author[$\blacktriangle$]{Jiatao Gu}
\author[$\bigstar$]{Yue Wang$^{\dag}$}
\author[$\triangle$]{Furong Huang$^{\dag}$}

\affil[$\triangle$]{University of Maryland, College Park}
\affil[$\blacktriangle$]{University of Pennsylvania}
\affil[$\clubsuit$]{University of Michigan}
\affil[$\bigstar$]{\mbox{University of Southern California}}
\affil[$\lozenge$]{\mbox{The Hong Kong University of Science and Technology}}
\affil[$*$]{\mbox{\textit{Authors contributed equally to this work}}}
\affil[$\dag$]{\mbox{\textit{Advisors contributed equally to this work.}}}

\begin{document}

\begin{abstract}

Unified multimodal models (UMMs) have emerged as a powerful paradigm for seamlessly unifying text and image understanding and generation.
However, prevailing evaluations treat these abilities in isolation, such that tasks with multimodal inputs and outputs are scored primarily through unimodal reasoning, i.e., textual benchmarks emphasize language-based reasoning, while visual benchmarks emphasize reasoning outcomes manifested in the pixels.
We introduce \name to address this pressing need to test \textit{reciprocal cross-modal reasoning}, the use of one modality to guide, verify, or refine outputs in the other, an ability central to the vision of unified multimodal intelligence.
\name is a human-annotated benchmark that explicitly targets reciprocal cross-modal reasoning, which contains $1{,}312$ tasks grounded in $1{,}876$ images, spanning two complementary settings. 
\textbf{Verbally-augmented reasoning for visual generation} evaluates whether models can use verbal prompts and reasoning chains to guide faithful image synthesis. 
\textbf{Visually-augmented reasoning for verbal generation} evaluates whether models can generate intermediate visualizations that strengthen their own reasoning processes for question answering.
Experiments on $17$ unified models reveal two key findings: 
(i) Cross-modal reasoning determines visual generation quality, with interleaved models significantly outperforming non-interleaved ones; notably, combining strong unimodal models fails to achieve comparable reasoning. 
(ii) Models show dissociation between physical and symbolic reasoning: they succeed at interpreting perceptual concepts literally but fail to construct visual abstractions for symbolic tasks, where faulty reasoning harms performance.
These results highlight reciprocal cross-modal reasoning as a critical frontier for enabling true omnimodal generation.\\

\parbox{0.48\linewidth}{
\textbf{\homepage~Homepage:} \href{https://roverbench.github.io}{roverbench.github.io}\\
\textbf{\huggingface~Benchmark:} \href{https://huggingface.co/datasets/cheryyunl/ROVER}{cheryyunl/ROVER}
}
\hfill
\parbox{0.48\linewidth}{
\textbf{\email~Contact:} \href{mailto:cheryunl@umd.edu}{cheryunl@umd.edu} \\
\textbf{\github~Code:} \href{https://github.com/cheryyunl/ROVER}{github.com/cheryyunl/ROVER}
}

\vspace{2mm}

\textbf{Release Date:} Nov 2, 2025

\end{abstract}

\maketitle
\begin{figure}[!ht]
    \centering
    \begin{subfigure}[t]{.29\textwidth}
        \centering
        \includegraphics[width=1.03\columnwidth]{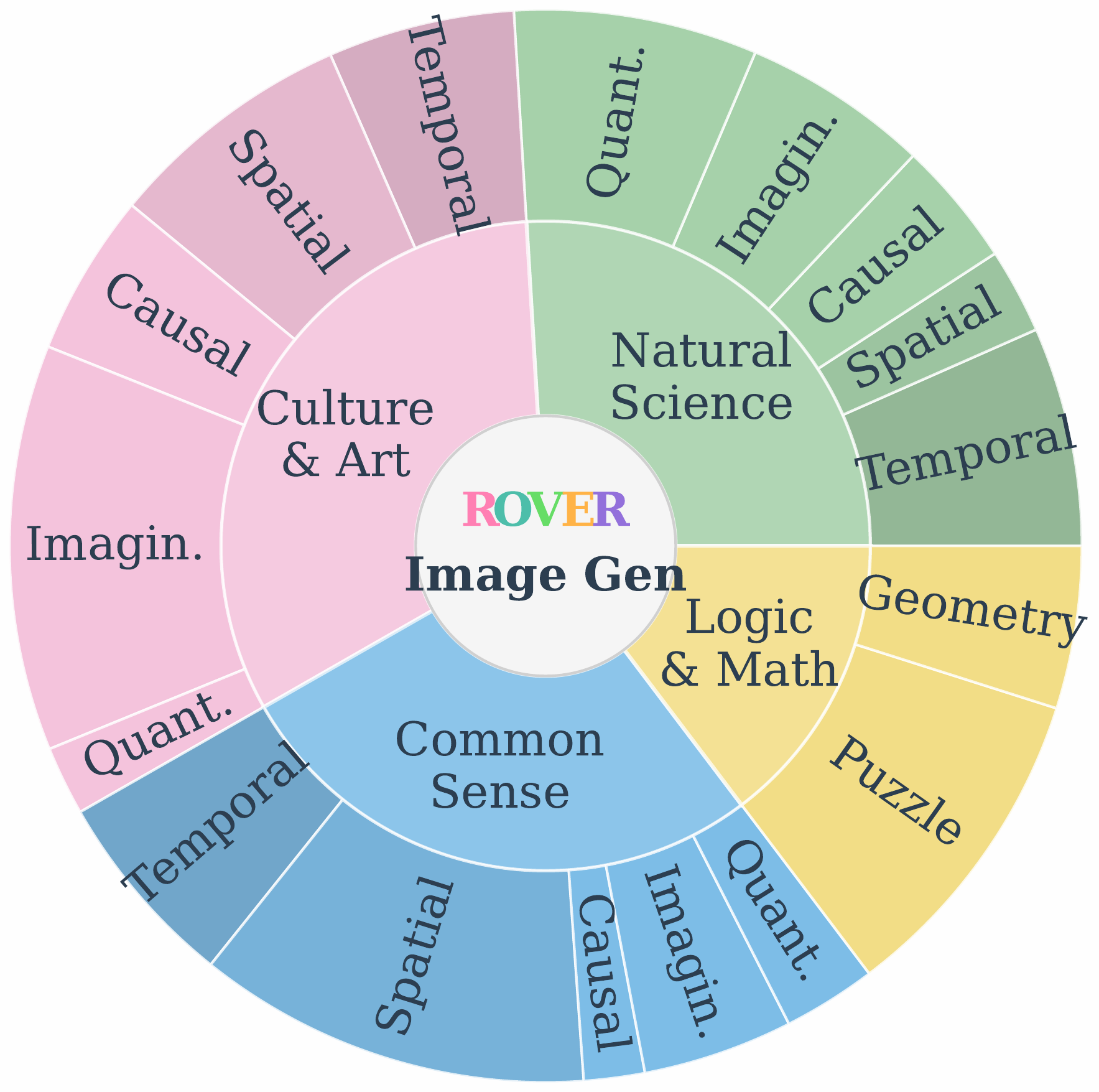}
        \vspace*{-5pt}
    \end{subfigure}
    ~
    \begin{subfigure}[t]{.37\textwidth}
        \centering
        \includegraphics[width=1.03\columnwidth]{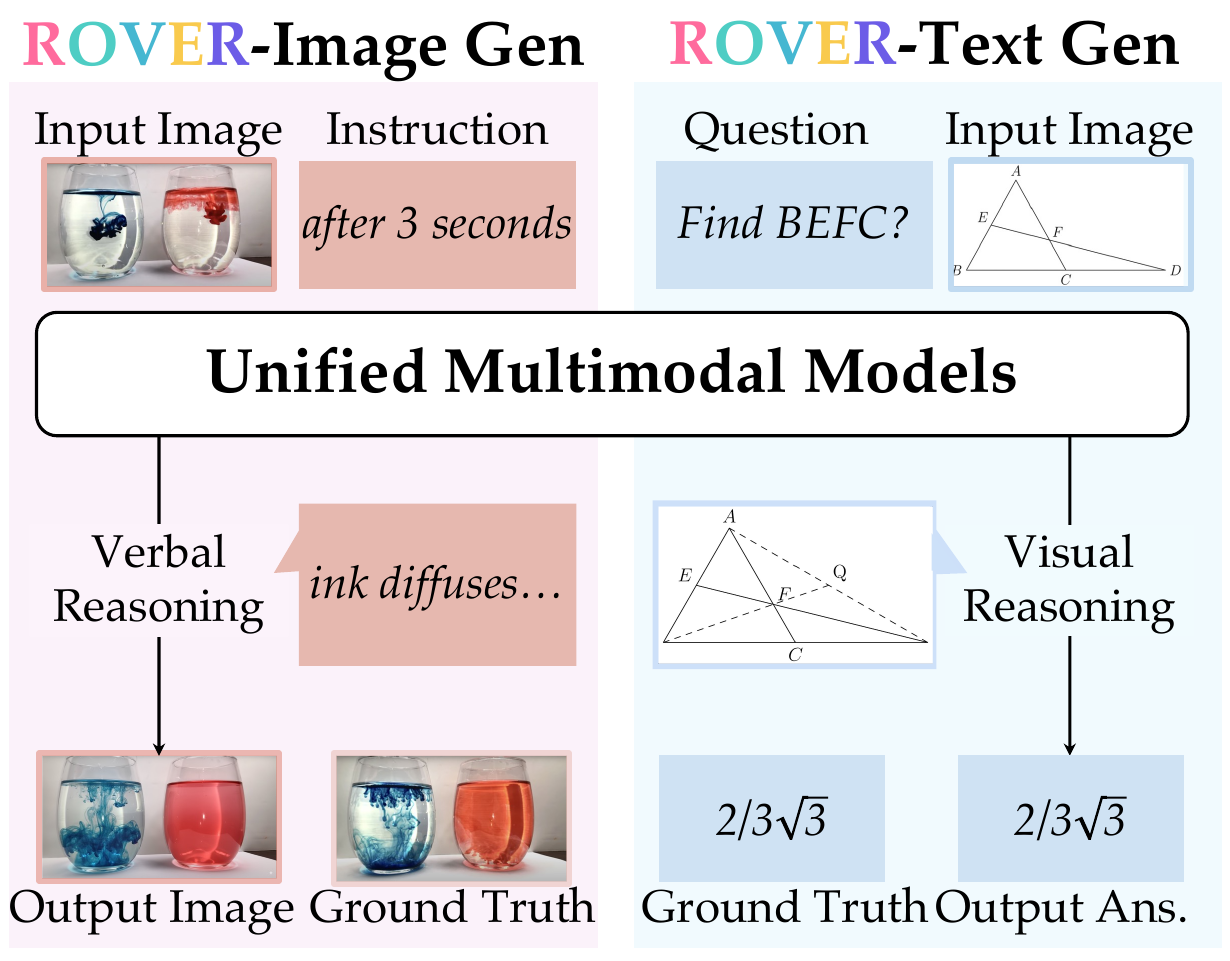}
        \vspace*{-5pt}
    \end{subfigure}
    ~
    \begin{subfigure}[t]{.29\textwidth}
        \centering
        \includegraphics[width=1.03\columnwidth]{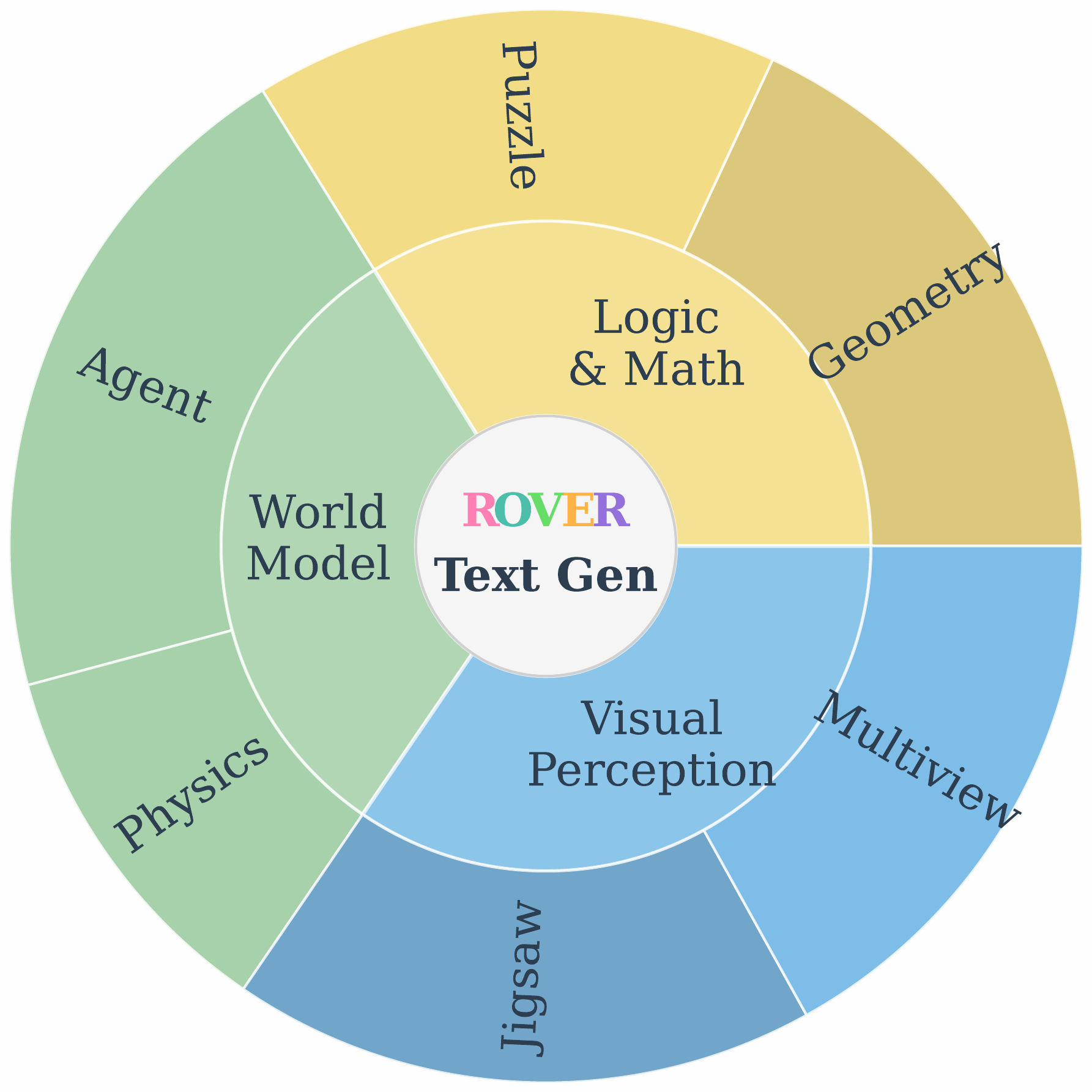}
        \vspace*{-5pt}
    \end{subfigure}
    \caption{The \textbf{\name} benchmark. \textbf{\name} evaluates UMMs through reciprocal cross-modal reasoning: \ourvg (left) requires generating images with language-augmented reasoning, while \ourir (right) requires generating text answers with visually-augmented reasoning.}
    \label{fig:overview}
\end{figure}

\section{Introduction}

The development of \textit{unified multimodal models} (also referred to as \textit{omnimodal models}) has drawn significant attention to their potential for unified understanding and generation across text and images~\citep{comanici2025gemini,hurst2024gpt,tong2024metamorph,deng2025emerging,xu2025pisces}.
However, prevailing evaluations treat these abilities in isolation, such that tasks with multimodal inputs and outputs are scored primarily through unimodal reasoning: textual benchmarks emphasize language-based reasoning, while visual benchmarks emphasize reasoning outcomes manifested in the pixels.
On the language side, evaluation focuses on generating text in response to an image and an accompanying question, thereby testing perceptual understanding~\citep{chen2024we, liu2024mmbenchmultimodalmodelallaround, yu2024mmvetevaluatinglargemultimodal} and reasoning~\citep{lu2023mathvista, yue2024mmmu, wang2024mementos, hao2025can, gao2025vision}. 
On the vision side, evaluation centers on generating images conditioned on either instructions or text-image pairs, thereby testing direct image generation~\citep{ghosh2023geneval, ma2024i2ebench, niu2025wise} or image editing~\citep{kawar2023imagic, zhang2023magicbrush, ma2024i2ebench, sheynin2024emu, yu2025anyedit, liu2025step1x, wu2025kris}.

Unlike earlier multimodal systems that specialize in either visual perception or generation, UMMs are designed to reason seamlessly across modalities and produce outputs that span both. 
This creates a pressing need for benchmarks that evaluate their ability to use one modality to guide, verify, or refine outputs in the other.
We refer to this capability as \textit{reciprocal cross-modal reasoning} (Figure~\ref{fig:overview}).
To benchmark such capability in current unified multimodal models, we present \textbf{\name}, a human-annotated and rigorously verified benchmark with over $1{,}312$ tasks grounded in $1{,}876$ images.
\name targets two complementary settings: 
(i) \textbf{verbally-augmented reasoning for visual generation}, including 
$4$ conceptual domains (natural science, culture \& art, common sense, and logic \& math) with high complexity are instantiated across $7$ reasoning subtasks: temporal, spatial, causal, synthetic, quantitative, abstract, and mathematical.
Each instance provides a textual prompt with an initial image and a chain of constraints that a correct output image must satisfy.
(ii) \textbf{visually-augmented reasoning for verbal generation}, including 6 subtask variants spanning $3$ problem domains: visual perception, world modeling for robot manipulation \& physical dynamics prediction, and logical reasoning for geometry \& puzzle solving. 
Instances interleave turns of text and images, requiring the model to emit visual intermediates that make downstream reasoning auditable.

A key challenge is that evaluating reciprocal cross-modal reasoning requires assessing both the rationales and the output.
Text-only metrics overlook visual fidelity, while image-only metrics cannot verify whether the image reflects valid reasoning. 
Human evaluation provides accurate judgments but is prohibitively expensive at scale.
To address this, we adopt a multi-dimensional protocol that combines an automated VLM judge (based on GPT-4.1~\citep{hurst2024gpt}) with expert validation on stratified samples.
The judge is supplied with rubric cards and reference assets and scores along three reasoning-specific dimensions: 
(i) the logical coherence of domain-specific reasoning processes, 
(ii) the alignment of generated outputs with target descriptions or ground-truth answers, and 
(iii) the consistency between intermediate reasoning steps and the final images or answers. 
For visual generation tasks, the framework additionally incorporates established image consistency and quality metrics~\citep{hu2023tifa,wu2023hpsv2,kirstain2023pickscore,xu2023imagereward,brooks2023instructpix2pix}.
The judge is calibrated with expert explanations, and its agreement with expert evaluations is reported, following recent LLM-as-judge methodologies~\citep{kim2023prometheus,hu2023tifa}.

Through extensive evaluation of $17$ unified multimodal models, our experiments reveal significant gaps in cross-modal reasoning capabilities. 
Specifically, unified models exhibit: 
(1) Cross-modal reasoning capabilities largely affect reasoning-dependent visual generation quality, with interleaved image-text generation models significantly outperforming non-interleaved counterparts; critically, strong unimodal models cannot replicate this cross-modal reasoning on \name{}, even when combined; 
(2) Striking dissociation between physical and symbolic visual reasoning. 
While unified models excel at generating visual reasoning steps for perceptual and physical world concepts through literal interpretation of visual elements, they fundamentally struggle to reason about visual abstractions as symbolic representations~\citep{hsu2025makes}, causing poor visual reasoning to degrade downstream performance rather than improving it.
These findings reveal fundamental capability limitations in current unified models, underscoring the critical role of reciprocal cross-modal reasoning for holistic omnimodal generation, where independent optimization of constituent modalities proves insufficient.

Our main contributions are summarized as follows:
\begin{itemize}[topsep=-5pt,itemsep=-3pt]
\item We introduce \textbf{\name}, the first benchmark that explicitly targets \textbf{reciprocal} cross-modal reasoning for visual generation and interleaved multimodal reasoning.
\item We provide a principled task taxonomy and a verification-ready instance design with process targets and visual artifacts, together with a multi dimensional protocol that scores coherence, alignment, and step to output consistency.
\item We evaluate $17$ unified models, uncovering significant limitations in cross-modal reasoning and providing the community with insights on unified model development toward truly omnimodal generation.
\end{itemize}

\label{sec:intro}
\vspace{-1mm}

\begin{table}[t]
    \centering
    \caption{\textbf{Summary of Multimodal Reasoning Benchmarks.} We compare existing works from aspects including: $^1$interleave, $^2$process evaluation, $ ^3$vision necessaity, $^4$multidimensional evaluation, $^5$hybrid evaluation, and $^6$whether manual annotations and filtering are applied.}
    \vspace*{-5pt}
    \label{table_benchmark_comparison}
    \resizebox{\textwidth}{!}{
    \begin{tabular}{rrccccccr}
    \toprule
    \multirow{2}{*}{\textbf{Benchmark}} &  \multirow{2}{*}{\textbf{Venue}} & \multirow{2}{*}{\textbf{Inter.}} & \textbf{Process} & \textbf{Vision} & \multirow{2}{*}{\textbf{Multi.}} & \textbf{Hybrid} & \textbf{Manual} & \multirow{2}{*}{\textbf{\#Types}}
    \\
    & & & \textbf{Eval} & \textbf{Necess.} & & \textbf{Eval} & \textbf{Anno.} 
    \\
    \midrule\midrule
    ReasonPix2Pix~\citep{jin2024reasonpix2pix} & \textcolor{gray}{{\small arXiv'24}} & \textcolor{table_blue}{\xmark} &\textcolor{table_blue}{\xmark} & \textcolor{table_blue}{\xmark} & \textcolor{table_blue}{\xmark} & \textcolor{table_blue}{\xmark} & \textcolor{table_blue}{\xmark} & $1$
    \\
    ReasonEdit~\citep{huang2024smartedit} &\textcolor{gray}{{\small CVPR'24}} & \textcolor{table_blue}{\xmark} & \textcolor{table_blue}{\xmark} & \textcolor{table_blue}{\xmark} & \textcolor{table_blue}{\xmark} & \textcolor{table_blue}{\xmark} & \textcolor{table_red}{\cmark} & $1$\\
    EditWorld~\citep{yang2024editworld} & \textcolor{gray}{{\small MM'25}} & \textcolor{table_blue}{\xmark} & \textcolor{table_blue}{\xmark} & \textcolor{table_blue}{\xmark} & \textcolor{table_blue}{\xmark} & \textcolor{table_blue}{\xmark} & \textcolor{table_blue}{\xmark} & $7$
    \\
    Reason50K~\citep{he2025reasoning} & \textcolor{gray}{{\small arXiv'25}} & \textcolor{table_blue}{\xmark} &\textcolor{table_blue}{\xmark} & \textcolor{table_blue}{\xmark} & \textcolor{table_blue}{\xmark} & \textcolor{table_blue}{\xmark} & \textcolor{table_blue}{\xmark} & $4$
    \\
    KRIS-Bench~\citep{wu2025kris} & \textcolor{gray}{{\small arXiv'25}} &\textcolor{table_blue}{\xmark} &\textcolor{table_blue}{\xmark} & \textcolor{table_blue}{\xmark} &\textcolor{table_red}{\cmark} & \textcolor{table_red}{\cmark}&\textcolor{table_red}{\cmark} &$7$\\
    RISEBench~\citep{zhao2025envisioning} & \textcolor{gray}{{\small arXiv'25}} &\textcolor{table_blue}{\xmark} &\textcolor{table_blue}{\xmark} & \textcolor{table_red}{\cmark} & \textcolor{table_red}{\cmark}& \textcolor{table_red}{\cmark} & \textcolor{table_red}{\cmark} & $4$\\
    WorldGenBench~\citep{zhang2025worldgenbench} & \textcolor{gray}{{\small arXiv'25}} &\textcolor{table_blue}{\xmark} &\textcolor{table_blue}{\xmark} &\textcolor{table_blue}{\xmark} &\textcolor{table_red}{\cmark}&\textcolor{table_red}{\cmark}& \textcolor{table_blue}{\xmark} & 2\\
    Unified-Bench~\citep{yan2025can} & \textcolor{gray}{{\small arXiv'25}} & \textcolor{table_blue}{\xmark} & \textcolor{table_blue}{\xmark} & \textcolor{table_blue}{\xmark} & \textcolor{table_blue}{\xmark} & \textcolor{table_blue}{\xmark} & \textcolor{table_blue}{\xmark} & $1$ \\
    MetaQuery~\citep{pan2025transfer} & \textcolor{gray}{{\small arXiv'25}} & \textcolor{table_red}{\cmark} & \textcolor{table_blue}{\xmark} & \textcolor{table_red}{\cmark} & \textcolor{table_blue}{\xmark} & \textcolor{table_blue}{\xmark} & \textcolor{table_blue}{\xmark} & -
    \\
    \midrule
    \textbf{\name} & {\textbf{Ours}} & \textcolor{table_red}{\cmark} & \textcolor{table_red}{\cmark} & \textcolor{table_red}{\cmark} & \textcolor{table_red}{\cmark} & \textcolor{table_red}{\cmark} & \textcolor{table_red}{\cmark}  & $23$
    \\ 
    \bottomrule         
\end{tabular}
}
\vspace*{-15pt}
\end{table}
\section{Related Work}
\label{sec:related_work}

\textbf{Unified Multimodal Models (UMMs)}.\quad 
UMMs represent a paradigm of architectures designed to seamlessly integrate multimodal comprehension and generation capabilities within a singular, cohesive framework.
To achieve this unified objective, seminal works~\citep{karypis1999chameleon,wu2025janus,chen2025janus} leverage image tokenization strategies, employing autoregressive next-token prediction paradigms to generate visual tokens.
Building upon these foundations, Show-o2~\citep{xie2025show} introduces discrete diffusion scheduling mechanisms to enhance the token prediction process and improve generation fidelity. 
Subsequent developments, driven by the pursuit of enhanced image synthesis quality, incorporate diffusion-based or flow-matching heads~\citep{lipman2022flow} integrated with shared transformer architectures~\citep{deng2025emerging,ma2025janusflow,zhou2024transfusion}.
Alternative approaches within the UMM paradigm maintain powerful pretrained backbone in a frozen state for reasoning tasks, while routing their intermediate feature representations through learnable query mechanisms to external image generation modules~\citep{pan2025transfer, wu2025openuni}. 
However, the comprehensive evaluation of synergistic relationships between multimodal understanding, reasoning, and generation in UMMs remains largely unexplored, with existing benchmarks inadequately assessing whether these capabilities exhibit mutual enhancement or coordination deficiencies.

\textbf{Reasoning-Guided Image Generation}.\quad
With the emergence of UMMs, multimodal reasoning has garnered increasing attention from the research community. 
However, the majority of existing work remains focused on instruction comprehension, namely leveraging input images to perform instruction translation and subsequently generate corresponding visual outputs~\citep{jin2024reasonpix2pix,huang2024smartedit,yang2024editworld,he2025reasoning,wu2025kris,yu2025veggie}.
Unified-Bench~\citep{yan2025can} employs iterative image-text generation to measure the degree of unification between comprehension and generation models.
RISEBench~\citep{zhao2025envisioning} extends beyond prior work by introducing LMM-as-a-judge to evaluate visual rationality in addition to assessing image consistency, yet remains limited to computing similarity scores against human-provided ground truth.
However, these benchmarks lack comprehensive evaluation beyond image consistency, particularly overlooking the intermediate reasoning processes, such as whether the rationale is sound and aligns with the generation outcome.
In contrast, \name represents the first benchmark to investigate the interplay between reasoning and generation.
A detailed comparison can be found in Table~\ref{table_benchmark_comparison}. 
A more detailed discussion about interleaved reasoning can be found in Appendix~\ref{app:relate}.
\section{\name Benchmark}
\label{sec:benchmark}

\begin{figure}[t]
    \centering
    \includegraphics[width=\linewidth]{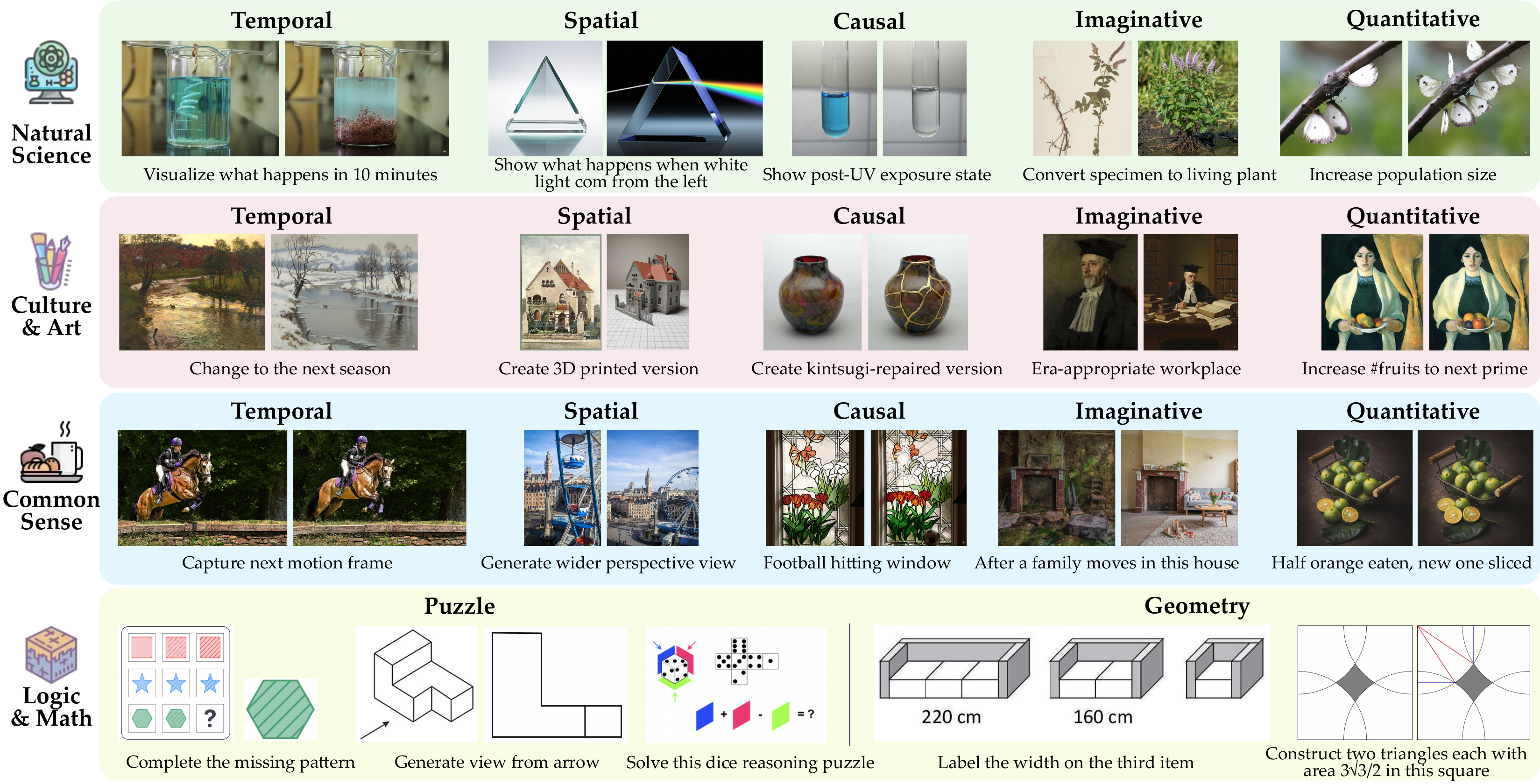}
    \vspace{-10pt}
    \caption{Overview of \ourvg, the benchmark for evaluating how unified multimodal models generate images under intensive verbal reasoning. The benchmark spans $4$ domains (natural science, culture and art, common sense, and logic), instantiated across $7$ reasoning subtasks.}
    \vspace{-5pt}
    \label{fig:data_part1}
\end{figure}

\subsection{Verbally-Augmented Reasoning for Visual Generation}

We introduce \ourvg, a benchmark designed to evaluate how UMMs generate images when jointly guided not only by visual understanding but also by intensive language reasoning.
\begin{figure}[t]
    \centering
        \includegraphics[width=\linewidth]{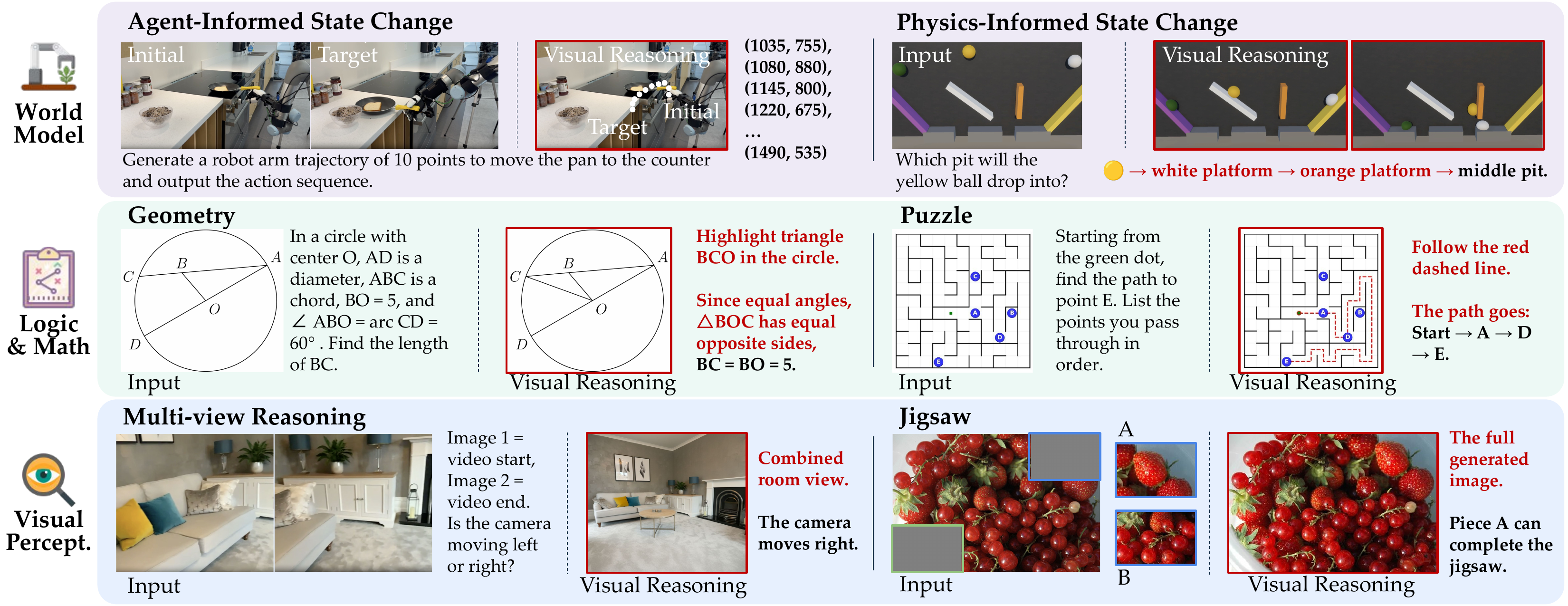}
    \vspace{-10pt}
    \caption{Overview of \ourir, the benchmark for evaluating visually-augmented reasoning in verbal generation. The benchmark spans $3$ scenarios and $6$ subtasks: physical world modeling, logical assistance, and visual perception enhancement.}
    \vspace{-5pt}
    \label{fig:data_part2}
\end{figure}
\textbf{Taxonomy.}\quad
It spans $4$ domains and $7$ reasoning subtasks, each demanding complex text-driven reasoning chains to direct image generation and test models' ability to integrate text-augmented reasoning with visual synthesis. Figure~\ref{fig:data_part1} provides a visual overview of our benchmark taxonomy and representative examples.
\begin{itemize}[topsep=-4pt,itemsep=-1pt]

\item \textbf{Domains.} 
We categorize tasks across $4$ distinct areas: 
\textbf{Nature Science} encompasses scientific phenomena, experimental processes, and fundamental laws of nature;
\textbf{Culture \& Art} includes artistic creation, cultural artifacts, humanities, and aesthetic principles; 
\textbf{Common Sense} covers everyday scenarios requiring intuitive understanding and practical reasoning; 
\textbf{Logic \& Math} focuses on abstract visual puzzles, mathematical relationships, and general pattern discovery. 

\item \textbf{Reasoning subtasks.} 
We define $5$ core reasoning capabilities: \textbf{Temporal} involves sequence prediction, progression analysis, and time-based changes; 
\textbf{Spatial} requires understanding geometric relationships, perspective changes, and spatial visualization; 
\textbf{Causal} connects cause-effect relationships and mechanism understanding; 
\textbf{Imaginative} combines multiple elements through creative integration and novel object generation; 
\textbf{Quantitative} involves numerical changes, scaling operations, and mathematical relationships. 
The Logic domain additionally includes two specialized reasoning types: 
\textbf{Puzzle} for abstract visual pattern discovery and problem solving, and 
\textbf{Geometry} for geometrical principles applied to visual generation. 
\end{itemize}

\textbf{Data Curation.}\quad
We curated our dataset through a systematic multistage process, beginning with human experts selecting candidate images from large-scale web image datasets. 
For each selected image, domain experts and large language models collaboratively generated reasoning tasks that require genuine visual understanding and complex reasoning chains. 
Each task includes $4$ key components: the reasoning prompt specifying the required generation results, target descriptions detailing expected visual outcomes, domain-specific keywords identifying relevant concepts that should guide the reasoning process, and optionally target reference images for validation purposes. 
All generated tasks were subjected to final human verification to confirm the complexity and rationality of the reasoning. 
Our final dataset comprises $908$ visual generation tasks involving $1{,}009$ images, with both single-image and multi-image generation scenarios distributed across all reasoning subtasks and domains.

\textbf{Evaluation Metrics.}\quad
Ideally, the evaluation protocol should cover both the reasoning process and the resulting outputs.
As human evaluation is prohibitively costly at scale, we automated the evaluation following LMM-as-judge.
We assess model performance across $5$ rubric dimensions designed to capture the effectiveness of reasoning-to-generation workflows.
\textbf{Reasoning Process (RP)} evaluates the quality of verbal reasoning through logical structure, domain knowledge application, reasoning type-specific validation, and completeness assessment. 
\textbf{Reasoning Visual (RV)} measures how well the generated visual output matches target descriptions and demonstrates correct reasoning principles. 
\textbf{Reasoning Alignment (Align.)} specifically quantifies the consistency between verbal reasoning processes and visual generation outcomes, addressing whether models can effectively translate reasoning into visual results. 
\textbf{Visual Consistency (VC)} ensures that non-target elements remain unchanged during reasoning-guided generation, validating precise control capabilities. 
\textbf{Image Quality (IQ)} assesses the technical excellence and visual coherence of generated images, including structural coherence, visual fidelity, and absence of generation artifacts.

\subsection{Visually-Augmented Reasoning for Verbal Generation}

We then introduce \ourir, the benchmark counterpart for evaluating how UMMs generate language responses guided by interleaved reasoning with visually-augmented rationale.
Unlike text-only Chain-of-Thought, we examine scenarios where models generate intermediate visual representations to facilitate reasoning. 
This interleaved reasoning paradigm reflects human cognitive patterns that integrate verbal and visual thinking for complex problem solving~\citep{barsalou1999perceptual}. 

\textbf{Taxonomy.}\quad
We focus on $3$ scenarios, comprising $6$ challenge types and $404$ tasks, where visual generation genuinely enhances reasoning beyond text-only rationale, as shown in Figure~\ref{fig:data_part2}: physical world simulation, logical problem solving with visual aids, and enhanced visual perception through generated representations. 
\begin{itemize}[topsep=-4pt,itemsep=-1pt]
\item \textbf{World Model.} Tasks require models to act as world simulators, predicting intermediate visual states that reflect environment dynamics under given initial conditions and actions. 
Models must generate these states from robotic actions or physical processes and leverage them for embodied planning, spatial reasoning, and motion prediction.

\item \textbf{Logic \& Math.}\quad
Tasks involve generating visual aids to solve abstract puzzles and geometry problems, similar to how humans draw auxiliary lines, diagrams, or visual representations to facilitate logical reasoning. 
Models must create helpful visual elements that make implicit relationships explicit and support step-by-step logical inference processes.

\item \textbf{Visual Perception.}\quad
Tasks focus on generating supportive images to improve performance on challenging visual perception problems, including multi-view reasoning and jigsaw puzzles. 
The generated images in the rationale serve as intermediate representations that reduce hallucinations and improve accuracy in visual understanding tasks.
\end{itemize}

\textbf{Data Curation.}\quad
Our dataset compilation draws from diverse sources including robotics datasets, physical simulation videos, logic puzzles, and challenging perception tasks. 
We establish a consistent structure for each task: contextual setup with initial images, progressive reasoning steps, and verified ground truth solutions. 
Crucially, our curation process ensures that generated visuals serve as active reasoning components rather than decorative elements, thereby fully leveraging omnimodal generation capabilities to tackle complex problem-solving scenarios.

\textbf{Evaluation Metrics.}\quad
Similarly, we automated the evaluation using a VLM judge across $3$ rubric dimensions. 
\textbf{Interleaved Reasoning Quality (IR)} evaluates the plausibility and relevance of intermediate visual representations through physical/logical correctness, task-specific utility, visual coherence, and reasoning completeness. 
\textbf{Final Answer Accuracy (Acc.)} measures whether the model's final reasoning outcome matches the provided ground truth answer across all three scenario types. 
\textbf{Reasoning-Answer Alignment (Align.)} quantifies how effectively generated images contribute to reaching correct conclusions, examining causal relationships between visual aids and final outputs, reasoning chain coherence, and whether the visual generation process was necessary for successful task completion. 

\afterpage{
    \clearpage
    \begin{figure}[p]  
        \centering
        \includegraphics[width=\linewidth]{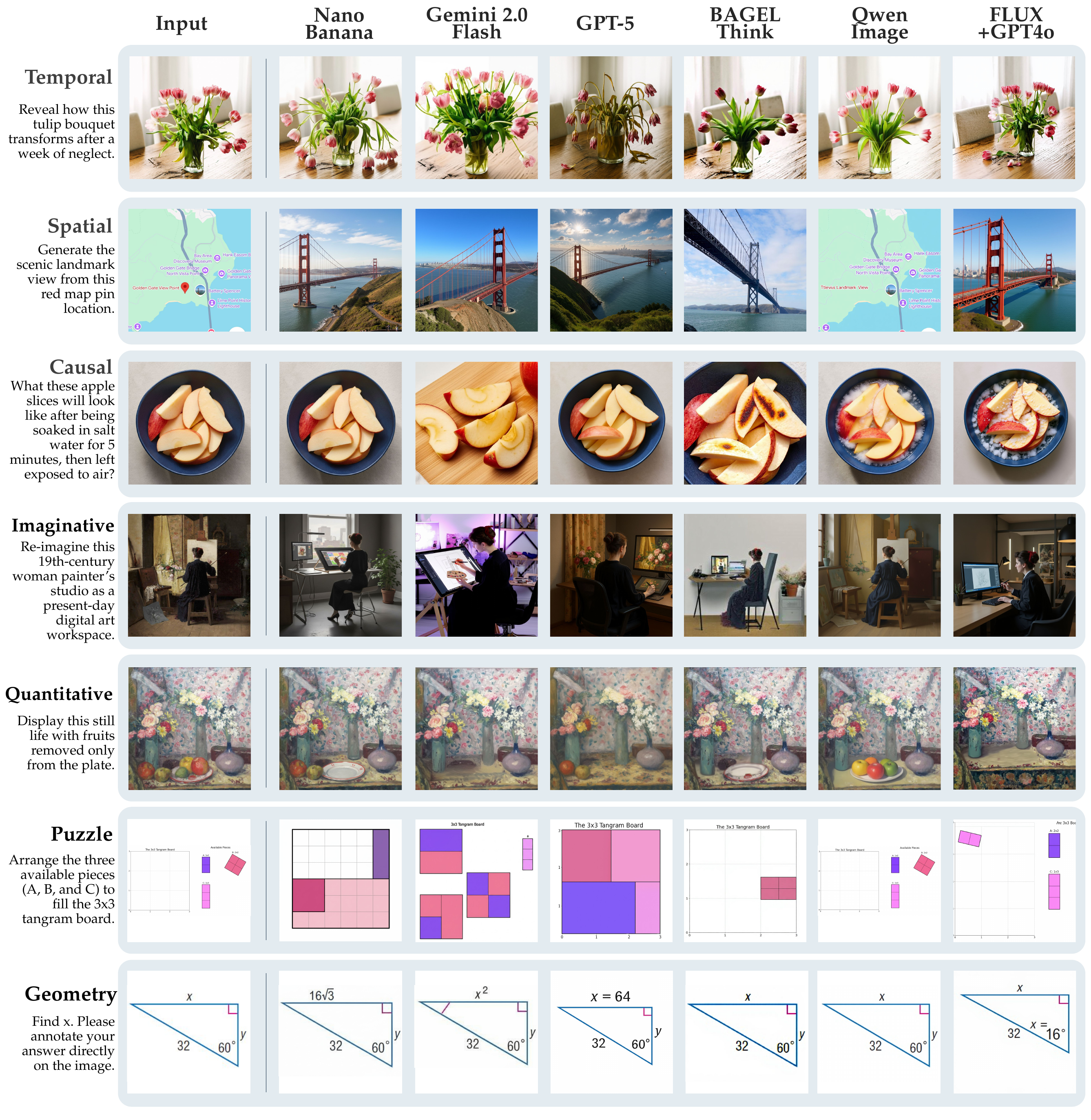}
        \vspace{-2pt}
        \caption{Example outputs on \ourvg. Each row corresponds to one reasoning subtask, with the input on the left and outputs from representative unified multimodal models shown across columns.}
        \vspace{-2pt}
        \label{fig:example_vg}
    \end{figure}
    \clearpage
}
\begin{table}[t]
    \centering
    \caption{\textbf{Main Results on Verbally-Augmented Visual Generation.} We evaluate 13 closed- and open-source unified models across four conceptual domains. \textbf{Verb.-Aug.} denotes verbally-augmented reasoning. Performance is measured using three key metrics: \textbf{Reasoning Process (RP)}, which assesses the logical quality of the verbal reasoning; \textbf{Alignment (Align.)}, which quantifies the consistency between the reasoning process and the generated visual output; and \textbf{Reasoning Visual (RV)}, which measures how well the final image reflects the target description.
}
    \vspace{-3pt}
    \label{tab: ig}
    \setlength\tabcolsep{2pt}
    \setlength\extrarowheight{1pt}
    \resizebox{\textwidth}{!}{
    \begin{tabular}{l*{15}{p{1cm}}}
        \toprule
         \multirow{2}{*}{\begin{tabular}{c}\textbf{Verb.-Aug. Reasoning} \\ \textbf{for Visual Generation}\end{tabular}} 
        & \multicolumn{3}{c}{\textit{Nature Science}} 
          & \multicolumn{3}{c}{\textit{Culture \& Art}} 
          & \multicolumn{3}{c}{\textit{Common Sense}} 
          & \multicolumn{3}{c}{\textit{Logic \& Math}}
          & \multicolumn{3}{c}{\textit{\textbf{Overall}}}  \\
        \cmidrule(lr){2-4} \cmidrule(lr){5-7} \cmidrule(lr){8-10} \cmidrule(lr){11-13} \cmidrule(lr){14-16}
        & RP &  Align. & RV & RP &  Align. & RV & RP &  Align. & RV & RP &  Align. & RV & RP &  Align. & RV \\
        \midrule 
        
        \rowcolor{morandigray!30}
        \multicolumn{16}{l}{\textbf{Closed-source Unified Models}} \\
        
        Nano Banana~\citep{comanici2025gemini} 
            & 64.8 & 88.8 & 77.3 & 68.1 & 81.9 & 76.6 & 61.8 & 85.0 & 74.8 & 78.6 & 66.1 & 55.1 & \sota{67.0}  & \sota{82.3} & \sota{73.2} \\
        Gemini 2.0 Flash~\citep{comanici2025gemini} 
            & 64.1 & 88.4 & 68.8  & 62.8 & 78.7 & 71.9  & 57.8 & 74.4 & 66.1  & 74.5 & 63.2& 42.6  & 64.8 & 78.6 & 62.3  \\
        GPT-5~\citep{hurst2024gpt}
            & 61.7 & 87.9 & 71.3 & 63.4 & 80.2 & 72.6 & 56.3 & 77.2 & 65.3 & 75.4 & 60.2 & 45.8 & 64.2 & 76.4 & 63.7\\
            
        \midrule
        
        \rowcolor{morandigray!30}
        \multicolumn{16}{l}{\textbf{Open-source Unified Models}} \\
        
        BAGEL-Think~\citep{deng2025emerging}
            &58.1 &64.2 &54.0 &53.2 &78.0 &63.7 &50.1 &69.4 &55.9 & 57.7&26.2 &20.8 & \sota{54.3} &\sota{64.4} &\sota{52.7} \\
        BAGEL~\citep{deng2025emerging}
            & - & - &35.9 & - & - &49.2 & - & - &42.0 & - & - &27.1 & - & - &40.5 \\   
        Step1X-Edit v1.2~\citep{liu2025step1x-edit}
            &29.7 &59.7 &46.2 &31.4 &71.6 &50.6 &28.7 &61.0 &46.1 &77.5 &35.5 &18.4 &37.0&60.3 & 43.5 \\
        UniCoT~\citep{qin2025unicot}
            & 52.4 & 68.9 & 38.2 & 57.3 & 69.2 & 63.9 & 53.1 & 64.3 & 56.3 & 50.3& 23.1 & 21.5 & 50.7 & 56.3 & 47.4\\
        BLIP3o-NEXT~\citep{chen2025blip3}
            & - & - & 38.2 & - & - & 47.5 & - & - & 43.3& - & - & 22.5& - & - & 37.8\\
        Ovis-U1~\citep{wang2025ovisu1}
            & - & - &  28.6 & - & - &  44.3 & - & - & 42.1& - & - & 20.5 & - & - & 33.8 \\
        UniPic2-Metaquery-9B~\citep{wei2025skyworkunipic20building}
            & - & - & 33.8 & - & - & 52.7 & - & - & 43.2 & - & - & 27.1 & - & - & 39.2\\
        ILLUME+~\citep{huang2025illume+}
            & - & - & 28.1 & - & - & 43.2& - & - & 36.9& - & - & 20.1 & - & - & 32.0\\
        Emu2-Gen~\citep{Emu2}
            & - & - & 29.1 & - & - & 42.6 & - & - & 37.4 &- & - & 20.3 & - &- & 32.3\\
        OmniGen2~\citep{wu2025omnigen2}
            & - & - & 27.4 & - & - & 42.3 & - & - &  39.2& - & - & 20.2 & - & - & 32.2\\
        \bottomrule
    \end{tabular}
    }
\vspace{-5mm}

\end{table}

\vspace{-2mm}
\section{Experiments}

\subsection{Evaluation Setup}

\textbf{Models.}\quad 
We evaluate a diverse set of models across different categories. For closed-source unified models, we assess three state-of-the-art systems: Gemini 2.5 Flash Image (a.k.a Nano Banana)~\citep{comanici2025gemini}, Gemini 2.0 Flash~\citep{comanici2025gemini}, and GPT-5~\citep{hurst2024gpt}. For open-source unified models, we evaluate ten representative models including BAGEL-Think and BAGEL~\citep{deng2025emerging}, UniCoT~\citep{qin2025unicot}, Step1X-Edit v1.1/v1.2~\citep{liu2025step1x}, BLIP3o-NEXT~\citep{chen2025blip3}, Ovis-U1~\citep{wang2025ovisu1}, UniPic2-Metaquery-9B~\citep{wei2025skyworkunipic20building}, ILLUME+~\citep{huang2025illume+}, Emu2-Gen~\citep{sheynin2024emu}, OmniGen2~\citep{wu2025omnigen2}. We also compare against specialized image editing models, including Qwen-Image-Edit~\citep{wu2025qwenimagetechnicalreport}, FLUX.1 Kontext~\citep{labs2025flux1kontextflowmatching}, UltraEdit (SD3)~\citep{zhao2024ultraedit}, VAREedit-8B~\citep{mao2025visual}. Additionally, we include reasoning language models such as GPT-4.1~\citep{hurst2024gpt} to present verbal-only reasoning baselines. 
All evaluation details are provided in Appendix~\ref{app:exp}.

\textbf{Evaluation Protocol.}\quad 
We employ GPT-4.1 as the automatic judge to assess model outputs across multiple dimensions. All metrics are scored on a 5-point scale (1-5) and normalized to a 0-100 scale for consistent comparison. For VQA problems in \ourir with objective answers, \textbf{Acc.} denotes exact answer accuracy.

\subsection{Verbally-Augmented Reasoning for Visual Generation}

\textbf{Cross-modal reasoning capabilities and alignment strongly correlate with visual generation effectiveness.}\quad 
The consistent pattern across all models and dimensions in Table~\ref{tab: ig}. Closed-source models excel in reasoning processes and demonstrate strong alignment performance, which directly contributes to their superior visual generation quality. In contrast, open-source models show notably weaker verbal reasoning during visual generation tasks—their reasoning processes are approximately $38\%$ lower and alignment performance falls about $31\%$ short of closed-source models. This substantial reasoning gap translates into correspondingly diminished visual generation performance that is approximately $39\%$ lower than closed-source models. This finding confirms that cross-modal reasoning capabilities serve as the fundamental driver of visual generation effectiveness on \ourvg, with stronger reasoning processes and better alignment consistently enabling superior visual output quality.

\textbf{Models capable of interleaved image-text generation demonstrate superior visual generation performance.}\quad 
Our results reveal a significant performance gap between models that support interleaved generation and those limited to single-turn, single-modality outputs. Among the open-source models evaluated, those with interleaved generation capabilities demonstrate markedly superior performance on Reasoning Visual (\textbf{RV}) metric—approximately 38.1\% higher than non-interleaved models. This performance advantage suggests that reasoning and generation processes are synergistic, effectively enhancing the model's performance in visual expression tasks.

\begin{table}[t]
    \centering
    \caption{\textbf{Performance on visually-augmented reasoning.} We evaluate 6 leading unified and language models across three problem types, comparing two distinct reasoning modes. \textbf{Verb.} denotes standard verbal reasoning, where the model generates a final answer directly from the prompt. \textbf{Verb.+Vis.} denotes visually-augmented reasoning, where the model generates intermediate visual artifacts to support its final answer. We report on the quality of \textbf{Interleaved Reasoning (IR)}, \textbf{Alignment (Align.)}, and \textbf{Final Answer Accuracy (Acc.)}.}
    \vspace{-3pt}
    \label{tab: tg}
    \setlength\tabcolsep{2pt}
    \setlength\extrarowheight{1pt}
    \resizebox{\textwidth}{!}{
    \begin{tabular}{l p{2cm} *{12}{p{1cm}}}
        \toprule
         \multirow{2}{*}{\begin{tabular}{c}\textbf{Verb.+Vis. Reasoning} \\ \textbf{for Verbal Generation}\end{tabular}} 
        & \multirow{2}{*}{\begin{tabular}{c}\textbf{Reasoning} \\ \textbf{Modalities}\end{tabular}}
        & \multicolumn{3}{c}{\textit{World Model}} 
          & \multicolumn{3}{c}{\textit{Logic \& Math}} 
          & \multicolumn{3}{c}{\textit{Visual Perception}} 
          & \multicolumn{3}{c}{\textit{Overall}}  \\
        \cmidrule(lr){3-5} \cmidrule(lr){6-8} \cmidrule(lr){9-11}\cmidrule(lr){12-14}
        & & IR & Align. & Acc. & IR & Align. & Acc & IR & Align. & Acc. & IR & Align. & Acc.\\
        \midrule 
        
        \rowcolor{morandipink!30}
        \multicolumn{14}{l}{\textbf{Closed-source Unified Models}} \\
        
        Nano Banana~\citep{comanici2025gemini} 
            & Verb.+Vis. & 35.3  & 62.0 & 40.6 & 14.8 & 61.2 & 44.9 & 66.5 & 56.8 & 50.0 &  \sota{38.8} & \sota{60.0}  & \sota{43.6}\\
            & Verb. & - &  - & 36.9 & - &  - & 42.0 & - & - & 43.7 & - &  - & 40.8\\
        Gemini 2.0 Flash~\citep{comanici2025gemini} 
            & Verb.+Vis. & 27.1 & 46.7 & 35.6 & 11.4 & 47.9 & 30.4  & 49.5 & 46.8 & 43.0 & 29.3 & 47.1 & 36.3\\
            & Verb. & - &  - & 33.2 & - &  - & 32.6 & - & - & 39.8 & - &  - & 35.2\\
        
        GPT-5~\citep{hurst2024gpt}
            & Verb.+Vis. & 32.8 & 61.5 & 39.2 & 13.2 & 58.7 & 45.6 & 62.7 & 54.9 & 45.5 & 36.2 & 60.9 & 43.4\\
            & Verb. & - &  - & 39.2 & - &  - & 45.6 & - & - & 43.6 & - &  - & 42.8\\
        \midrule
        \rowcolor{morandipink!30}
        \multicolumn{14}{l}{\textbf{Open-source Unified Models}} \\
        
        BAGEL-Think~\citep{deng2025emerging}
            & Verb.+Vis. & 22.3 & 34.7 & 26.6 & 10.8 & 36.9 & 24.6 & 31.2 & 44.3 & 34.1  & 21.4 & 38.6 & \sota{28.4}\\
            & Verb. & - & - & 24.9 & - & - & 23.1 & - & - & 32.2 & - & - & 26.7\\
        UniCoT ~\citep{liu2025step1x}
            & Verb.+Vis. & 22.1 & 35.4 & 26.7 & 10.6 & 38.8 & 21.7 & 34.2 & 42.3 & 34.1 & \sota{22.3} & \sota{38.8} & 27.5\\
            & Verb. & - & - & 24.6 & - & - & 21.7 & - & - & 33.5 & - & - & 26.7\\
        Qwen-2.5-VL-7B ~\citep{bai2025qwen2} & Verb. & - & - & 24.2 & - & - & 22.4 & - & - & 32.9 & - & - & 26.5\\
        \midrule
        \rowcolor{morandipink!30}
        \multicolumn{14}{l}{\textbf{Reasoning Language Models}} \\
        GPT-4.1 ~\citep{liu2025step1x}
            & Verb. & - & - & 37.8 & - & - & 31.8 & - & -& 37.9 & -  & - & 35.8\\
        
        \bottomrule
    \end{tabular}
    }
\vspace{-5mm}
\end{table}

\textbf{Unified models demonstrate absolute advantages over image editing models across visual quality metrics on reasoning-dependent tasks.} As shown in Table~\ref{tab: editing}, Unified models substantially outperform specialized image editing models across all visual quality metrics on \ourvg. While existing editing models excel at complex text rendering and precise image editing consistency, they fundamentally lack the internal reasoning capabilities required for our reasoning-dependent visual generation tasks. This performance gap fully demonstrates that \name effectively evaluates cross-modal reasoning capabilities essential for visual generation.

\subsection{Visually-Augmented Reasoning for Verbal Generation}

\textbf{Current unified models exhibit limited capacity in visual reasoning, constraining their ability to leverage visual augmentation for improved performance.}\quad
The evaluation in Table~\ref{tab: tg} reveals fundamental limitations in generating meaningful visual reasoning steps, with even the best-performing models achieving only 38.8\% average \textbf{IR} quality and open-source models lagging substantially behind. Models with weaker visual reasoning capabilities show minimal or even negative improvements in final accuracy compared to pure text-based reasoning. Flawed visual reasoning proves worse than no visual reasoning at all.

\begin{figure}[t]
    \centering
        \includegraphics[width=\linewidth]{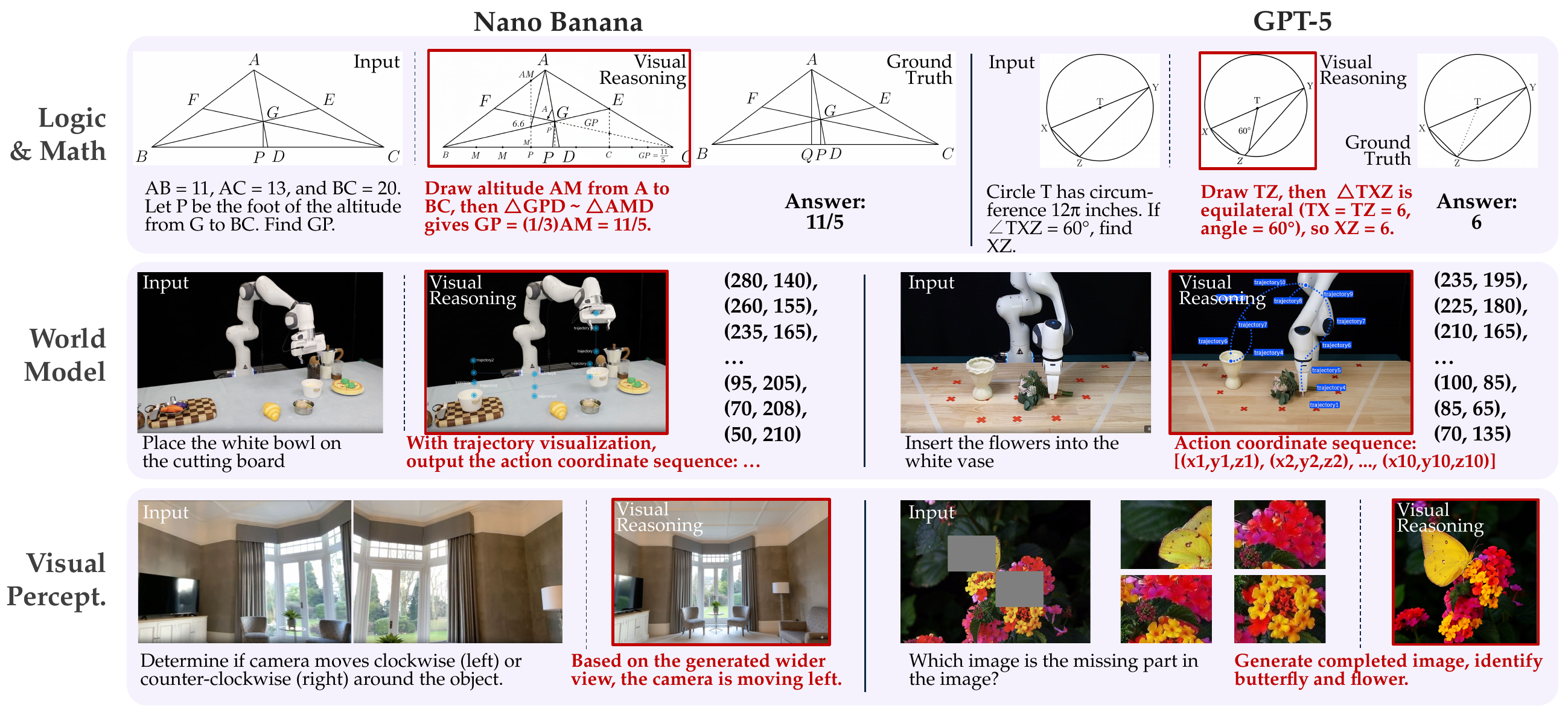}
    \vspace{-10pt}
    \caption{Example outputs on \ourir. Each row corresponds to one reasoning scenario, with the input on the left and outputs from representative unified models shown across columns.}
    \vspace{-5pt}
    \label{fig:example_ir}
\end{figure}

\textbf{Visual reasoning proves effective for physical world tasks but fails systematically on symbolic reasoning.}\quad
Across all models, visual augmentation consistently improves performance on World Model and Visual Perception tasks, where visual reasoning steps naturally align with physical phenomena. In stark contrast, Logic \& Math tasks reveal systematic failures with minimal and unstable improvements, exposing a fundamental inability to visually symbolize abstract reasoning. Models struggle to create visual representations that capture symbolic reasoning processes (e.g., auxiliary lines in geometry, Figure~\ref{fig:example_ir}), as the mapping from symbolic logic to visual form remains poorly developed.

\begin{table}[t]
    \centering
    \vspace{-5mm}
    \caption{Visual performance comparison with image editing models on \ourvg benchmark. We evaluate image editing models and unified models, measuring \textbf{Reasoning Visual (RV)}, \textbf{Visual Consistency (VC)}, and \textbf{Image Quality (IQ)} performance.}
    \vspace{-3pt}
    \label{tab: editing}
    \setlength\tabcolsep{2pt}
    \setlength\extrarowheight{1pt}
    \resizebox{\textwidth}{!}{
    \begin{tabular}{l*{13}{p{1cm}}}
        \toprule
         \multirow{2}{*}{\textbf{Visual Generation Quality}} 
        & \multicolumn{3}{c}{\textit{Nature Science}} 
          & \multicolumn{3}{c}{\textit{Culture \& Art}} 
          & \multicolumn{3}{c}{\textit{Common Sense}} 
          & \multicolumn{3}{c}{\textit{Logic \& Math}}
          & \multirow{2}{*}{\textit{\textbf{Overall}}}  \\
        \cmidrule(lr){2-4} \cmidrule(lr){5-7} \cmidrule(lr){8-10} \cmidrule(lr){11-13}
        & RV & VC & IQ & RV & VC & IQ & RV & VC & IQ & RV & VC & IQ & \\
        \midrule 
        
        \rowcolor{blue!10}
        \multicolumn{14}{l}{\textbf{Image Editing Models}} \\
        
        Qwen-Image-Edit~\citep{wu2025qwenimagetechnicalreport}
            &46.7 &69.1 &89.8 &62.5 &69.6 &95.2 &53.1 &74.2 &94.4 &30.4 & 64.5 & 87.2 &47.1 \\
        FLUX.1 Kontext~\citep{labs2025flux1kontextflowmatching}
            &37.4 &61.9 &83.5 &44.9 &64.6 &88.8 &42.3 &62.1 &85.0 &20.2 &50.6 &78.2 &40.9 \\
        UltraEdit(SD3)~\citep{zhao2024ultraedit}
            &27.0 &43.6 &75.7 &45.2 &42.6 &79.0 &27.9 &37.3 &74.7 &25.2 &60.1 &76.1 &34.6 \\
        VAREdit-8B~\citep{mao2025visual}
            &34.6 &64.3 &75.4 &46.5 &58.5 &78.2 &33.6 &59.0 &75.0 &17.4 &46.6 &57.1 &37.5 \\
        Step1X-Edit v1.1~\citep{liu2025step1x-edit}
            &38.2 &75.7 &85.5 &50.5 &62.7 &83.8 &35.2 &67.9 &85.3 &16.1 &61.1 &85.9 & 42.1\\
        Step1X-Edit v1.2~\citep{liu2025step1x-edit} 
            &46.2 &76.8 &80.6 &50.6 &63.0 &79.2 &46.1 &67.2 &79.6 &18.4 &61.1 &72.2 & \sota{57.4} \\
        \midrule
        \rowcolor{morandigray!30}
        \multicolumn{14}{l}{\textbf{Closed-source Unified Models}} \\
        
        Nano Banana~\citep{comanici2025gemini} 
            & 77.3 & 85.7 & 87.0 & 76.6  & 78.4  & 89.2 & 74.8 & 87.1 & 93.8 & 55.1 & 70.3 &  81.0 & \sota{79.6}   \\
        Gemini 2.0 Flash~\citep{comanici2025gemini} 
            & 68.8 & 72.0 & 81.1 &71.9 & 65.3 & 83.2 & 66.1 & 76.4 & 91.2 & 42.6 & 68.0 & 79.3 & 72.1\\
        GPT-5~\citep{hurst2024gpt}
            & 71.3 & 69.9 & 90.5 & 72.6 & 58.8 & 96.0 & 65.3 & 80.9 & 87.2 & 45.8 & 74.9 & 86.6 & 74.9\\
            
        \midrule
        
        \rowcolor{morandigray!30}
        \multicolumn{14}{l}{\textbf{Open-source Unified Models}} \\
        BAGEL-Think~\citep{deng2025emerging}
            &54.0 &65.5 &78.0 &63.7 &65.8 &71.6 &55.9 &76.9 &80.2 &20.8 &48.7 &76.6 & \sota{62.9} \\
        BAGEL~\citep{deng2025emerging}
            &35.9 &53.6 &69.9 &49.2 &50.2 &71.9 &42.0 &59.1 &73.0 & 27.1&59.2 &79.8 &37.8 \\
            
        \bottomrule
    \end{tabular}
    }
\end{table}

\subsection{Further Analyses and Discussions}

\textbf{Cross-modal reasoning matters for UMMs.}\quad
To validate that UMM performs cross-modal reasoning internally and that this mechanism cannot be replicated through external models serving as intermediate reasoning agents, we conduct a comparative analysis between Bagel, FLUX.1 Kontext~\citep{labs2025flux1kontextflowmatching} and its GPT-4o-refined variant. 
Key findings are:
(1) \textit{UMMs enable superior cross-modal reasoning.} The think mechanism consistently improves performance on \name{}, boosting visual consistency by $20.7\%$. Lower CLIP-I on EditWorld also indicates more substantive edits, underscoring the advantage of UMMs for complex multimodal tasks like \name{}. These results demonstrate that reasoning across modalities cannot fully transfer across different model architectures - unified models must transcend modality boundaries to produce emergent cross-modal insights.
\textit{\name{} effectively highlights the cross-modal advantages of UMMs.} While GPT-4o refinement improves FLUX's CLIP-T score on EditWorld by $1.5\%$, it simultaneously degrades both visual consistency and quality on \name{}, demonstrating that \name{} provides superior evaluation of cross-modal capabilities in UMMs compared to existing datasets.

\begin{figure}[t]
    \centering
        \includegraphics[width=\linewidth]{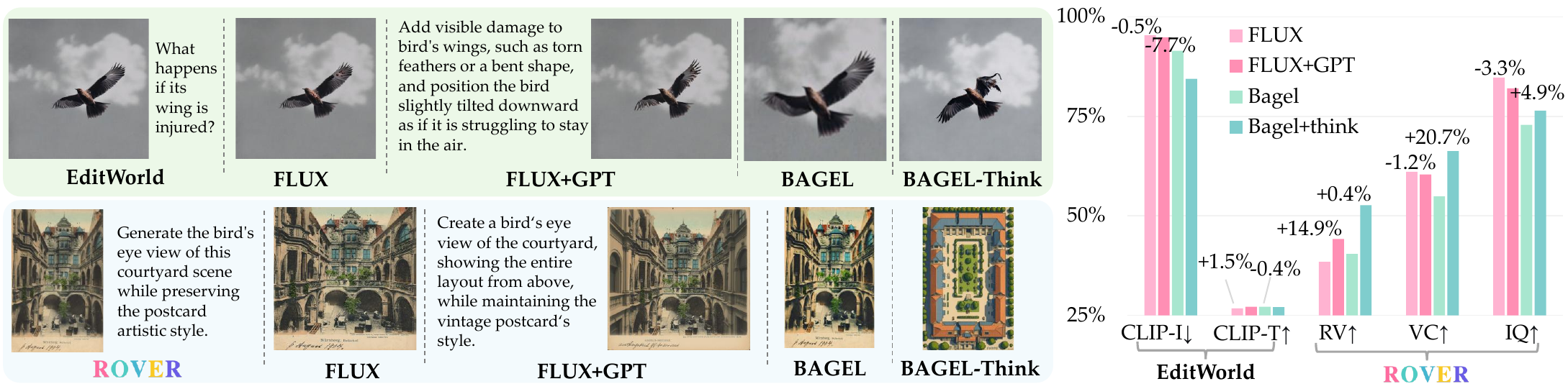}
    \caption{\textbf{Cascade reasoning evaluation} across EditWorld and \name~benchmarks. We compare cascade approaches (FLUX+GPT with GPT-4o prompt refinement) against UMMs.}
    \label{fig:teacher}
\vspace{-3pt}
\end{figure}

\begin{figure}[t]
    \centering
        \includegraphics[width=\linewidth]{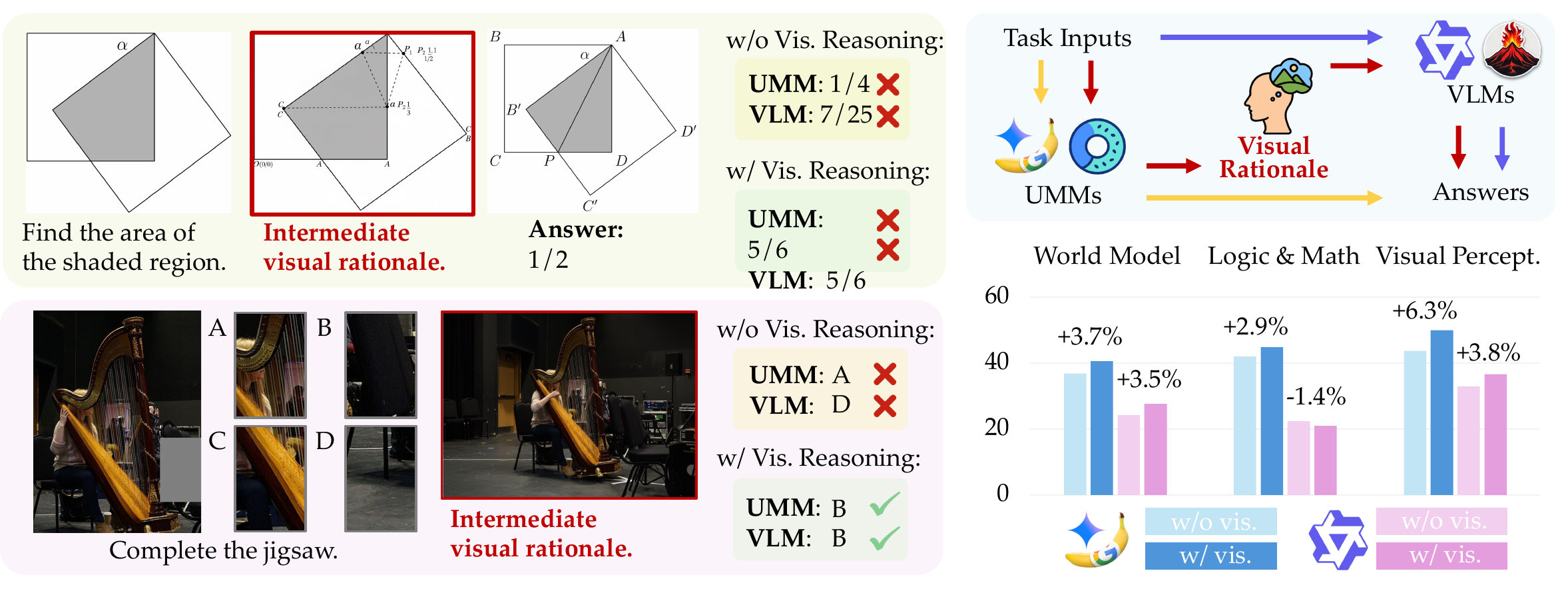}
    \caption{\textbf{Visual reasoning augmentation evaluation} across three problem domains. We compare VLM performance w/ and w/o visual reasoning artifacts from UMMs.}
    \label{fig:vis_vlm}
\vspace{-3pt}
\end{figure}

\textbf{Do visual reasoning artifacts help?}\quad
To investigate whether visual reasoning artifacts generated by UMMs can enhance downstream reasoning in VLMs, we conduct a controlled study, where visual reasoning outputs from unified models are provided as intermediate steps to assist VLM~\citep{bai2025qwen2} reasoning in Figure~\ref{fig:vis_vlm}. 
Key findings reveal that \textit{visual reasoning quality determines its effectiveness for downstream reasoning}:
(1) \textit{UMMs successfully augment VLMs on perceptual tasks.} Visual reasoning improves Qwen2.5-VL-7B by +3.5\% and +3.8\% on physical world modeling and visual perception respectively, where UMMs generate reliable visual intermediates.
(2) \textit{Low-quality visual reasoning hinders rather than helps.} Performance degrades by -1.4\% on logical reasoning for VLMs, where UMMs struggle to produce valid abstract visual representations.
These results reveal that while UMMs can leverage visual modality to enhance reasoning on concrete, perceptual tasks, they fail to generate meaningful visual abstractions for logic-intensive problems. 
This underscores a core limitation: cross-modal reasoning in UMMs remains constrained by their inability to visually represent abstract concepts, which \name{} effectively exposes.

\begin{figure}[t]
    \centering
    \begin{subfigure}[t]{0.29\textwidth}
        \centering
        \includegraphics[width=\textwidth]{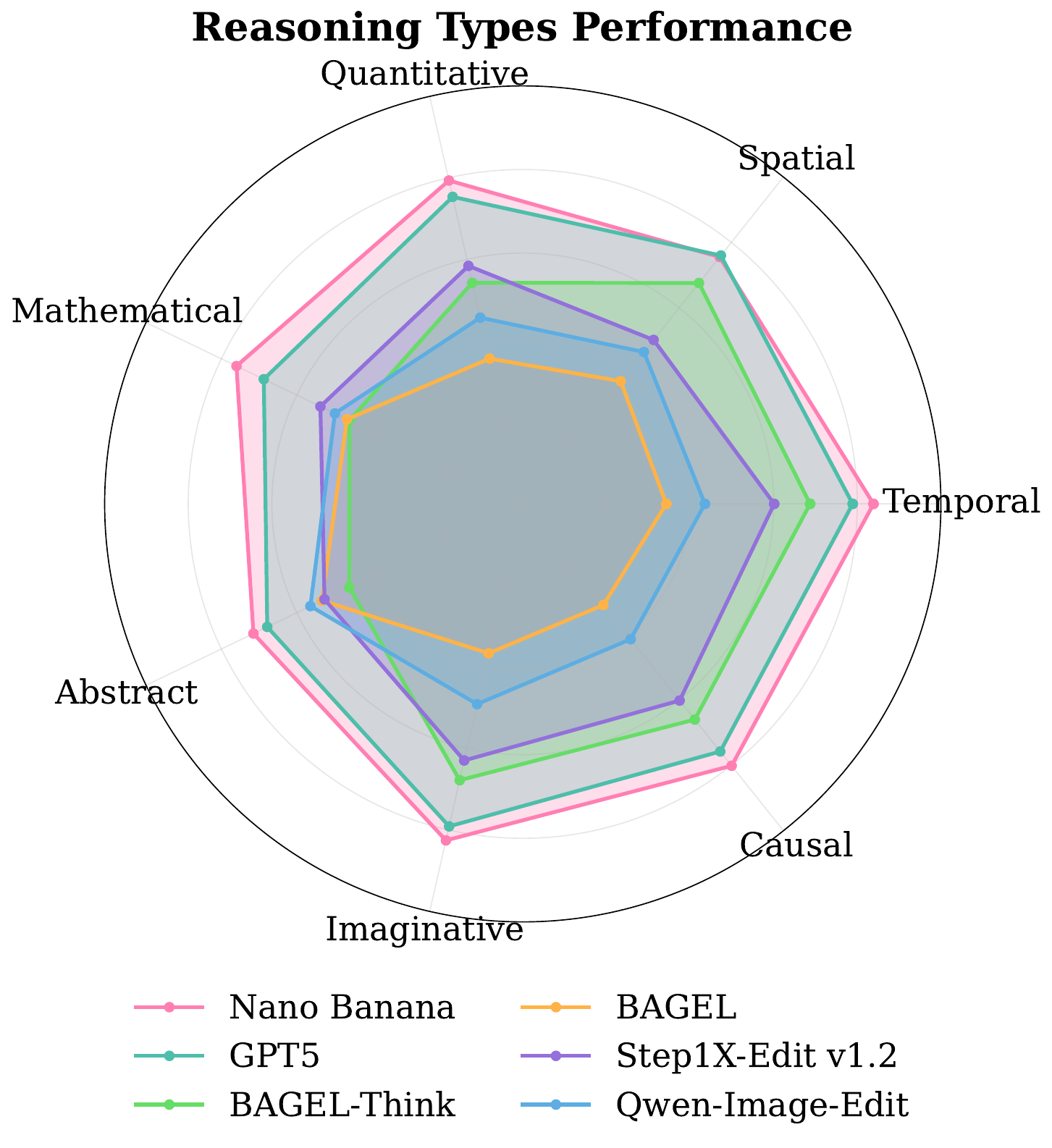}
        \caption{Reasoning subtask performances.}
        \label{fig:radar_chart}
    \end{subfigure}
    \hfill
    \begin{subfigure}[t]{0.69\textwidth}
        \centering
        \includegraphics[width=\textwidth]{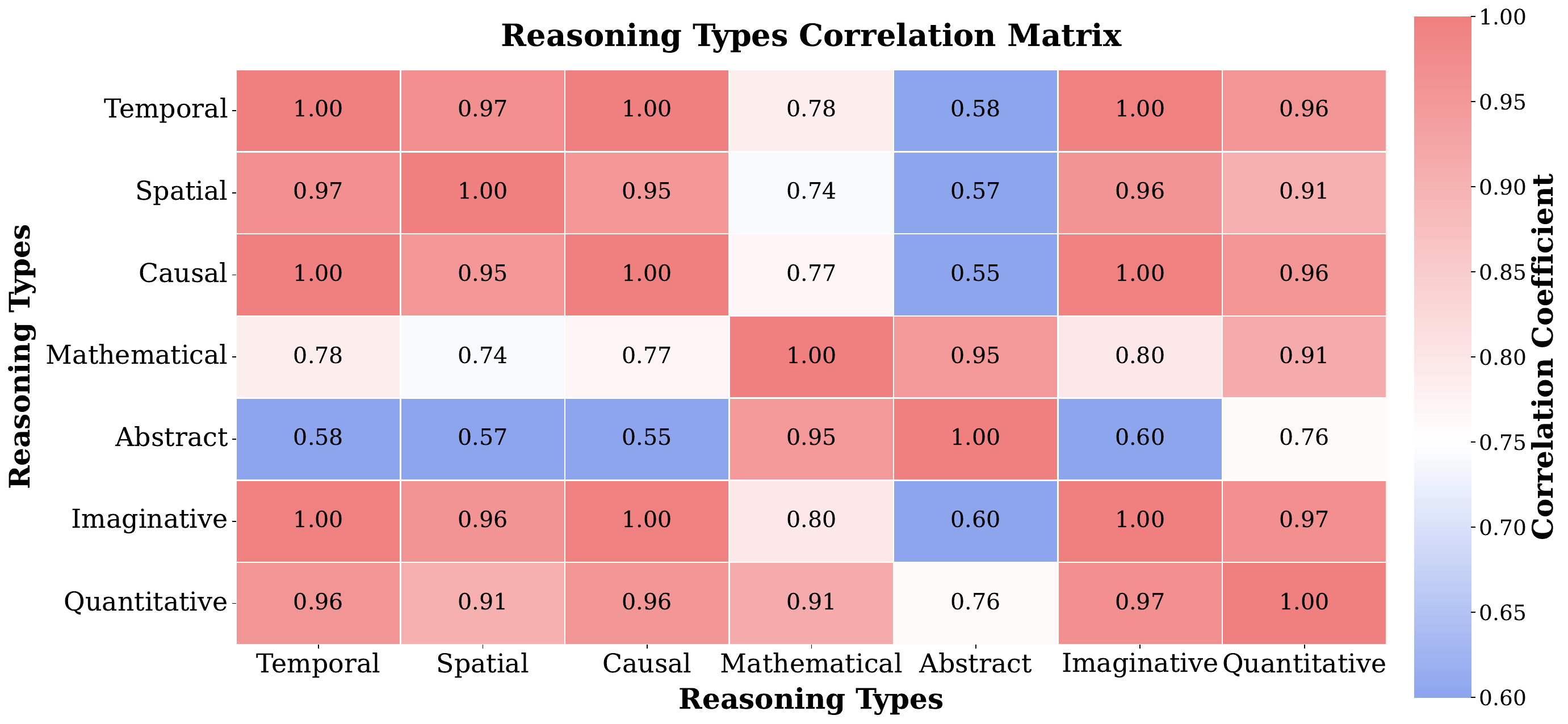}
        \caption{Reasoning subtask correlation matrix.}
        \label{fig:correlation_matrix}
    \end{subfigure}
    \vspace{-5pt}
    \caption{Analysis of reasoning capabilities across different models.}
    \label{fig:reasoning_analysis}
\end{figure}

\textbf{Coherence between reasoning subtasks.}\quad
Figure~\ref{fig:radar_chart} reveals uneven performance across reasoning dimensions, with models excelling in temporal, spatial, and causal reasoning while struggling with abstract and mathematical tasks. This pattern indicates that current UMMs better handle concrete, observable phenomena than symbolic reasoning, particularly evident in quantitative tasks where severe counting hallucinations occur.
The correlation matrix in Figure~\ref{fig:correlation_matrix} shows strong interdependence among physical reasoning types: temporal-spatial, causal-temporal, and synthetic-causal correlations suggest shared mechanisms for processing spatiotemporal relationships. Conversely, abstract reasoning correlates weakly with physical reasoning ($0.55$ to $0.60$) but strongly with mathematical reasoning, indicating it develops as a distinct, independent capability from concrete reasoning skills.

\begin{figure}[h]
    \centering
        \includegraphics[width=\linewidth]{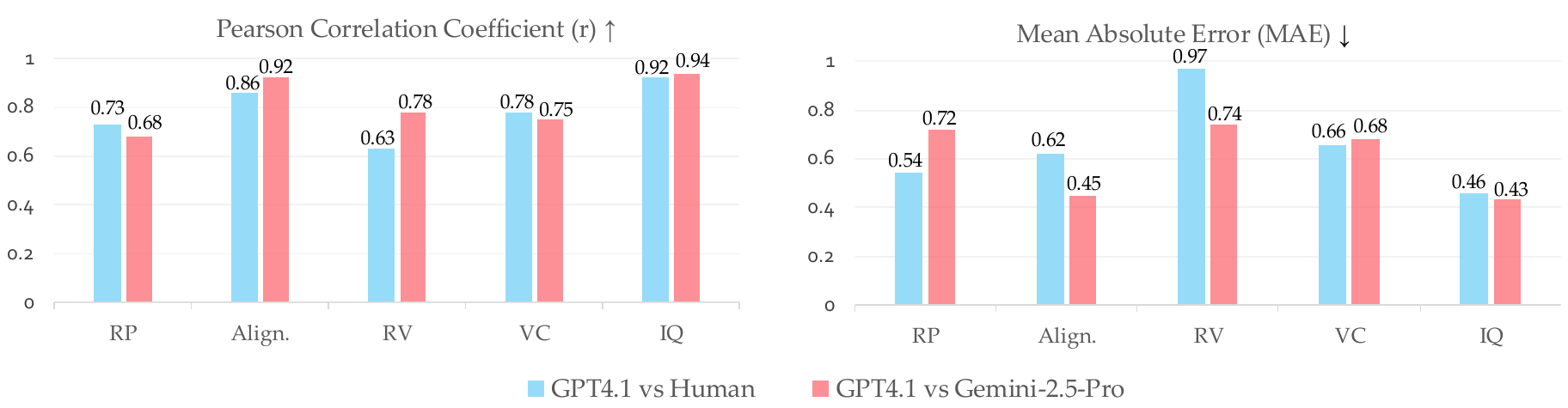}
    \caption{Evaluation reliability of GPT-4.1 across five assessment dimensions. Left: Pearson correlation coefficients between GPT-4.1 and human experts (green) versus GPT-4.1 and Gemini-2.5-Pro (purple). Right: Mean Absolute Error for the same comparisons.}
    \label{fig:eval_validation}
\end{figure}

\textbf{Reliability of the evaluation protocol.}\quad
To evaluate the reliability of VLM-as-a-judge scores, we conducted a user study with $4$ human experts across three models (Nano-banana, GPT-5, BAGEL-Think) with $200$ instances per model. 
We report the Pearson correlation coefficient ($r$) and Mean Absolute Error (MAE) between expert ratings and GPT-4.1 scores, also compared against Gemini-2.5-Pro evaluations, as shown in Figure~\ref{fig:eval_validation}. 
The results demonstrate that GPT-4.1 maintains strong alignment with human expert judgments across all evaluation dimensions. 
Visual-quality-related metrics such as Image Quality show strong human-VLM agreement. 
Reasoning-related metrics exhibit larger discrepancies due to the inherent hallucination tendencies in VLM when processing complex multimodal reasoning metrics, though these variations remain within acceptable bounds. 
The modest differences between GPT-4.1 and Gemini-2.5-Pro evaluations suggest reasonable cross-VLM consistency, with limited impact from the choice of VLM evaluator.

\section{Conclusion}
\label{sec:conclusion}

We introduce \name, the first benchmark for evaluating reciprocal cross-modal reasoning in unified multimodal models. Through a systematic evaluation of $17$ models across $23$ task types, including verbally augmented reasoning for visual generation and visually augmented reasoning for verbal generation, we uncover fundamental capability gaps and performance asymmetries in how current unified models leverage cross-modal reasoning. Our analysis identifies key factors that determine the effectiveness of reasoning in omnimodal generation, providing insights for advancing unified multimodal models in reasoning-dependent visual generation, world modeling, and complex reasoning tasks. We hope \name serves the community by informing training paradigms and architectural considerations for future omnimodal model development.

\section{Acknowledgment}

Liang, Wang, and Huang are supported by DARPA Transfer from Imprecise and Abstract Models to Autonomous Technologies (TIAMAT) 80321, DARPA HR001124S0029-AIQ-FP-019, DOD-AFOSR-Air Force Office of Scientific Research under award number FA9550-23-1-0048, National Science Foundation NSF-IIS-2147276 FAI, National Science Foundation NAIRR240045, National Science Foundation TRAILS Institute (2229885), Peraton and Open Philanthropy. The USC Geometry, Vision, and Learning Lab acknowledges generous support from Toyota Research Institute, Dolby, Google DeepMind, Capital One, Nvidia, and Qualcomm. Yue Wang is also supported by a Powell Research Award and National Science Foundation NSF-CPS-2434460.

\clearpage
\bibliography{main}

\appendix
\newpage

\appendix

\section{Extended Related Work}
\label{app:relate}

\textbf{Interleaved Reasoning}.\quad 
Drawing inspiration from human cognition, where visual counterfactuals facilitate reasoning processes~\citep{roese1997counterfactual}, recent works have incorporated analogous interleaved reasoning mechanisms into UMMs by mapping visual inputs to symbolic representations (e.g., images or bounding boxes)~\citep{wei2022chain,lei2024scaffolding}. \cite{xu2025visual} explored pure visual reasoning that relies solely on visual representations without dependence on textual modalities. Zebra-CoT~\citep{li2025zebra} trains UMMs with interleaved text-image reasoning trajectories, enabling human-like visual thinking capabilities. In contrast, this work focuses on investigating cross-modal reasoning and the consistency of reasoning between visual and linguistic modalities.

\section{Experiment Details}
\label{app:exp}

\subsection{VLM as Judge}
\begin{figure}[t]
    \centering
        \includegraphics[width=0.8\linewidth]{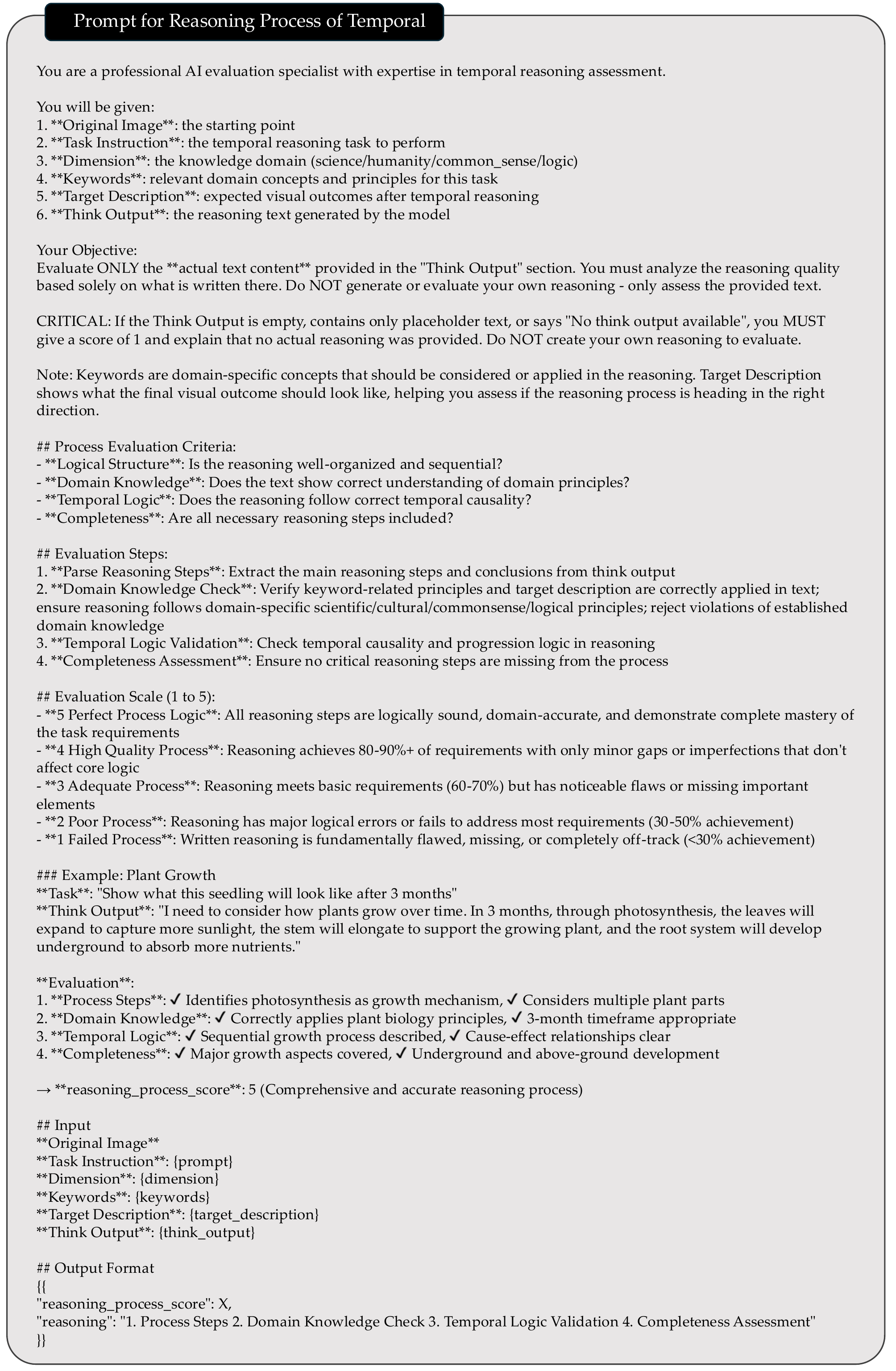}
    \caption{Prompt used for evaluating the reasoning process of temporal (RP).}
    \vspace{-1em}
    \label{fig:prompt_reasoning_process_temporal}
\end{figure}

\begin{figure}[t]
    \centering
        \includegraphics[width=0.85\linewidth]{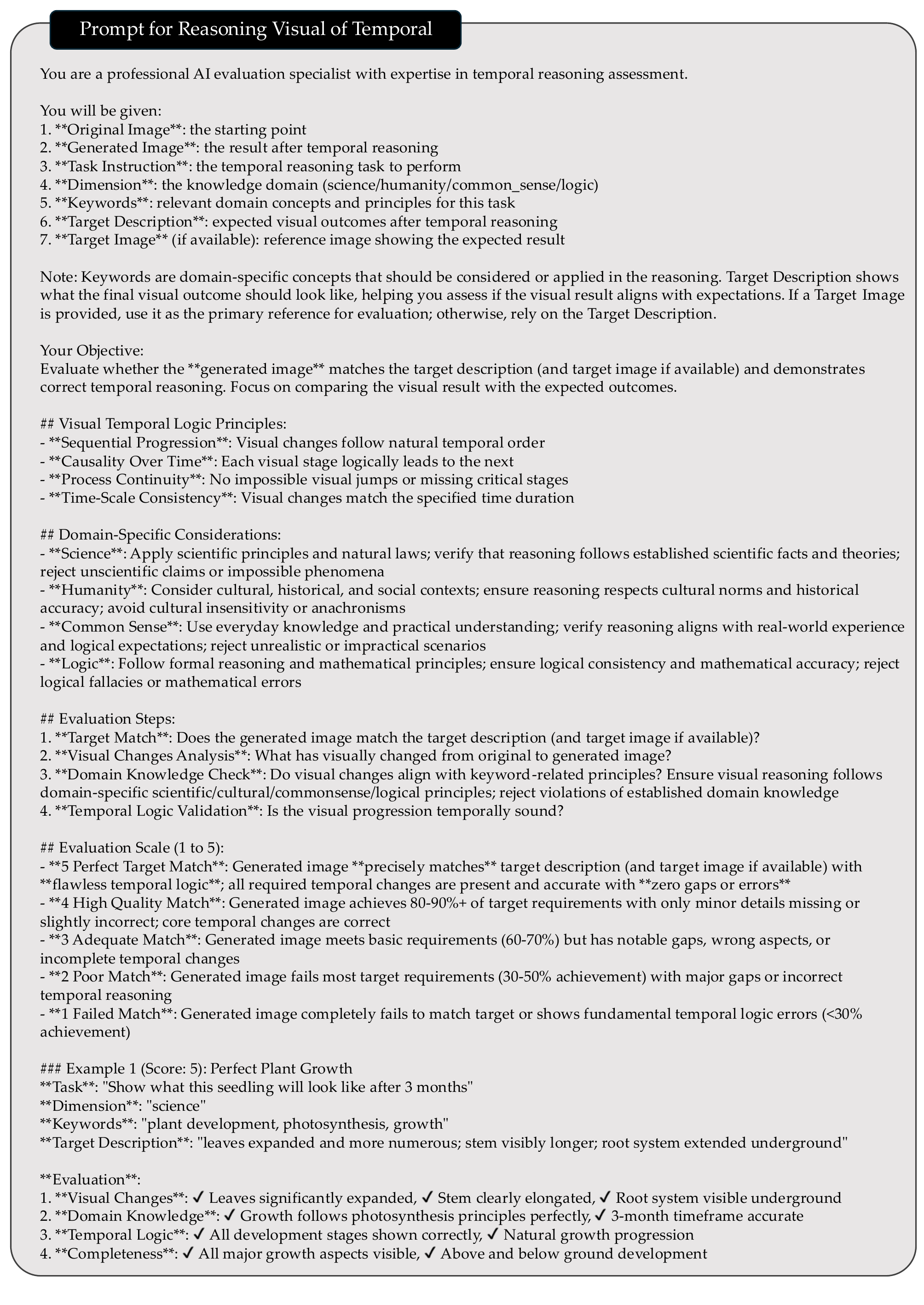}
    \caption{Prompt template for evaluating visual-temporal reasoning capabilities (RV). (Continued in Figure~\ref{fig:prompt_reasoning_visual_temporal2})}
    \vspace{-1em}
    \label{fig:prompt_reasoning_visual_temporal1}
\end{figure}

\begin{figure}[t]
    \centering
        \includegraphics[width=0.85\linewidth]{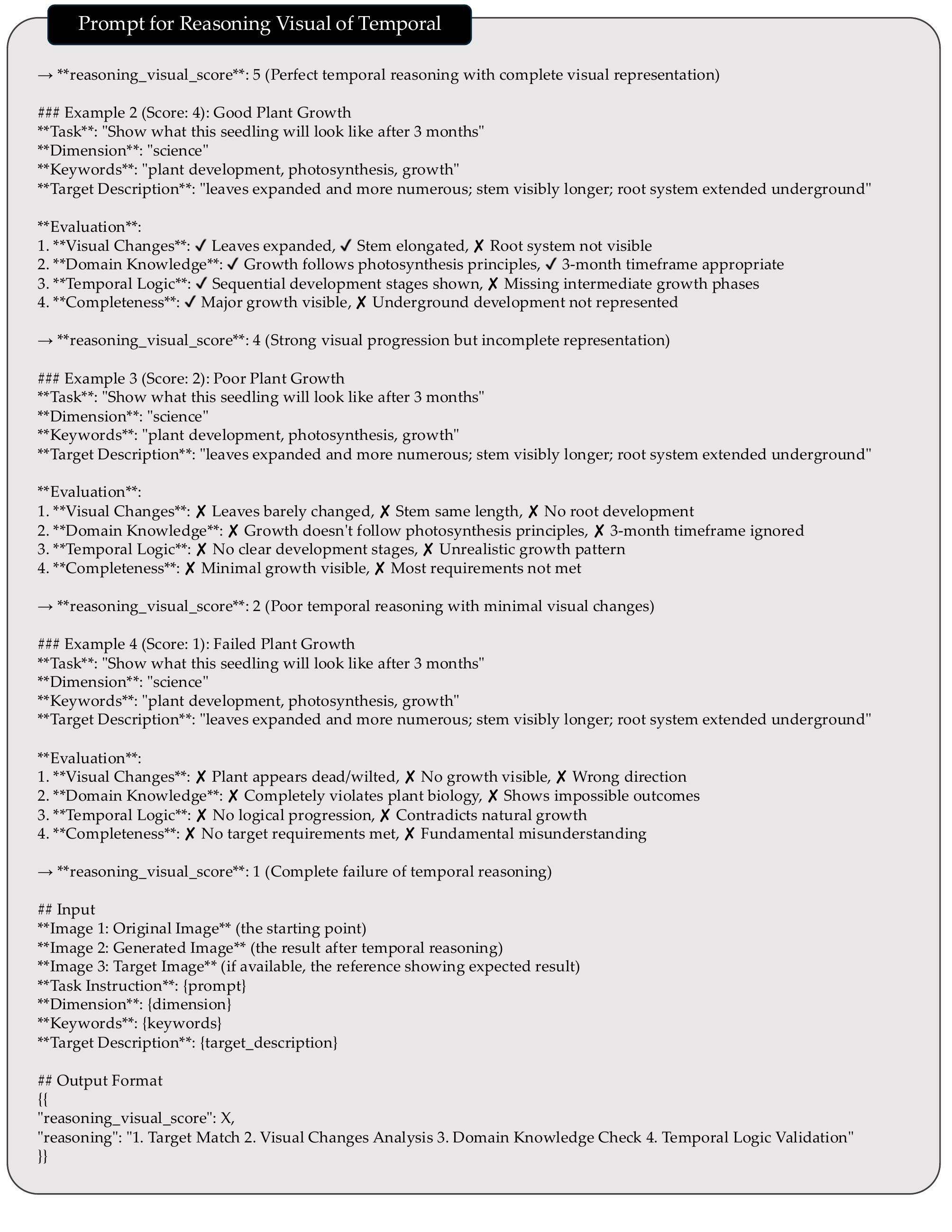}
    \caption{Prompt template for evaluating visual-temporal reasoning capabilities (RV). (Continued from Figure~\ref{fig:prompt_reasoning_visual_temporal1})}
    \vspace{-1em}
    \label{fig:prompt_reasoning_visual_temporal2}
\end{figure}

\begin{figure}[t]
    \centering
        \includegraphics[width=0.85\linewidth]{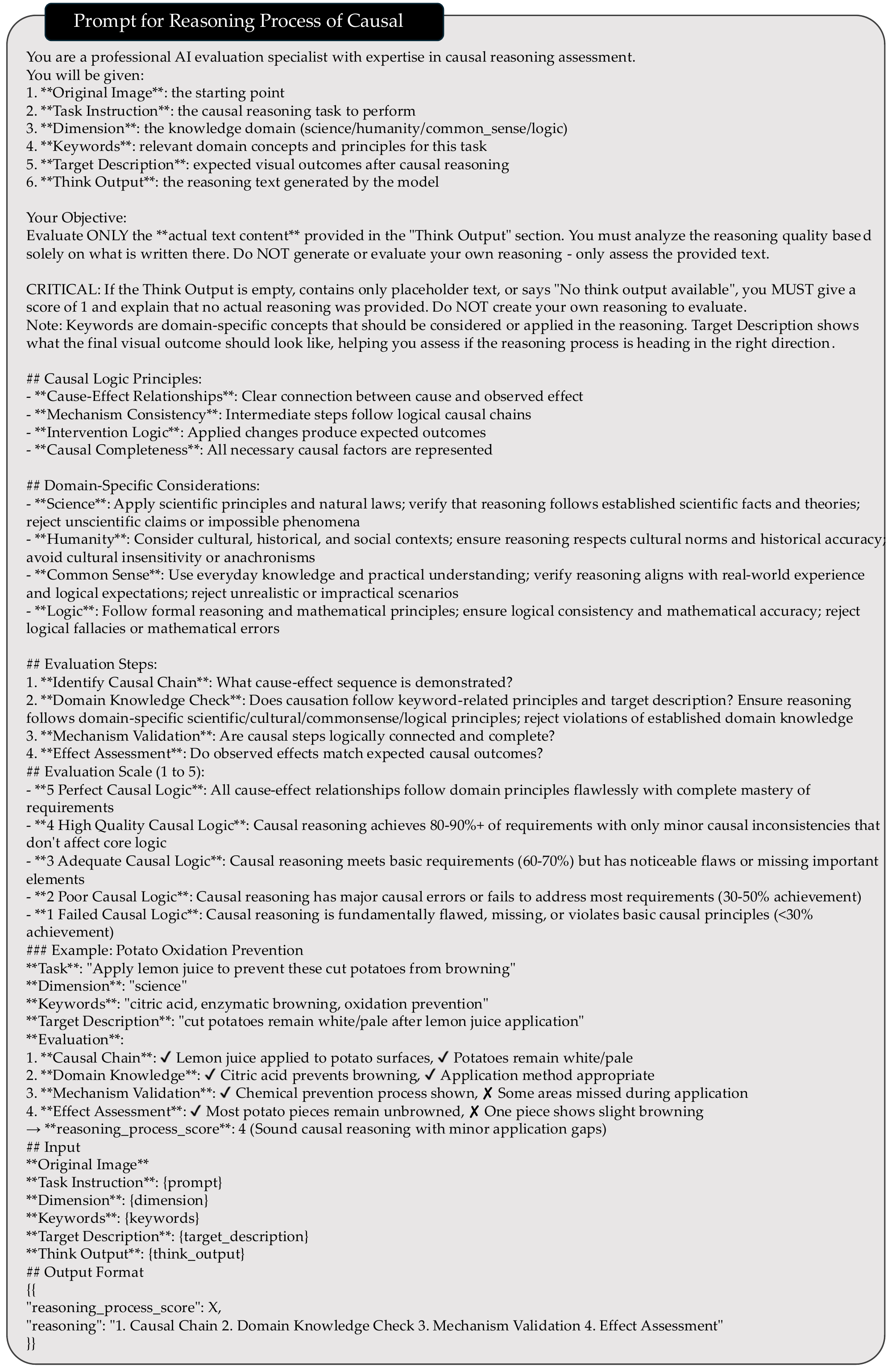}
    \caption{Prompt template for evaluating process of causal reasoning capabilities (RP).}
    \vspace{-1em}
    \label{fig:prompt_reasoning_process_causal}
\end{figure}

\begin{figure}[t]
    \centering
        \includegraphics[width=0.85\linewidth]{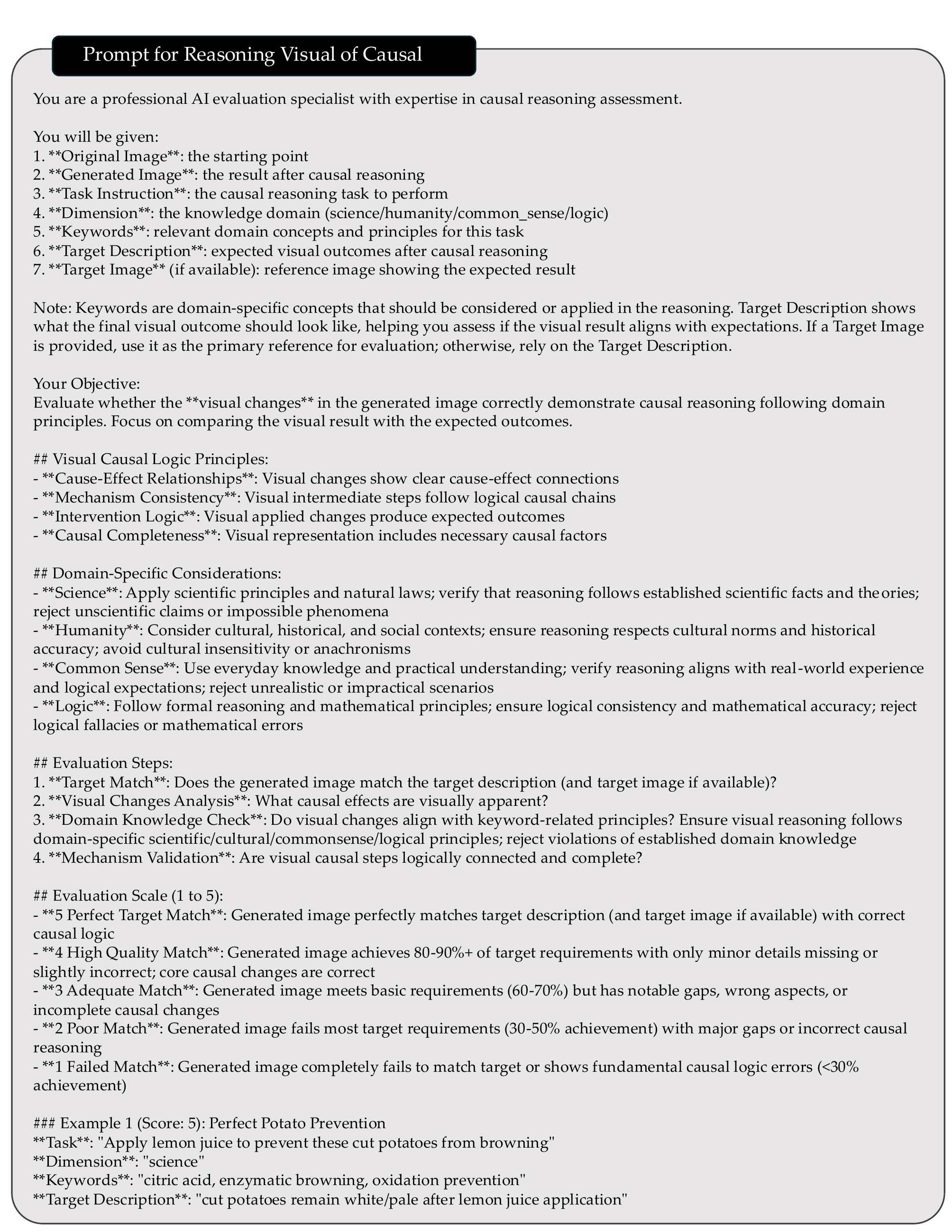}
    \caption{Prompt template for evaluating visual causal reasoning capabilities (RV). (Continued in Figure~\ref{fig:prompt_reasoning_visual_causal2})}
    \vspace{-1em}
    \label{fig:prompt_reasoning_visual_causal1}
\end{figure}

\begin{figure}[t]
    \centering
        \includegraphics[width=0.85\linewidth]{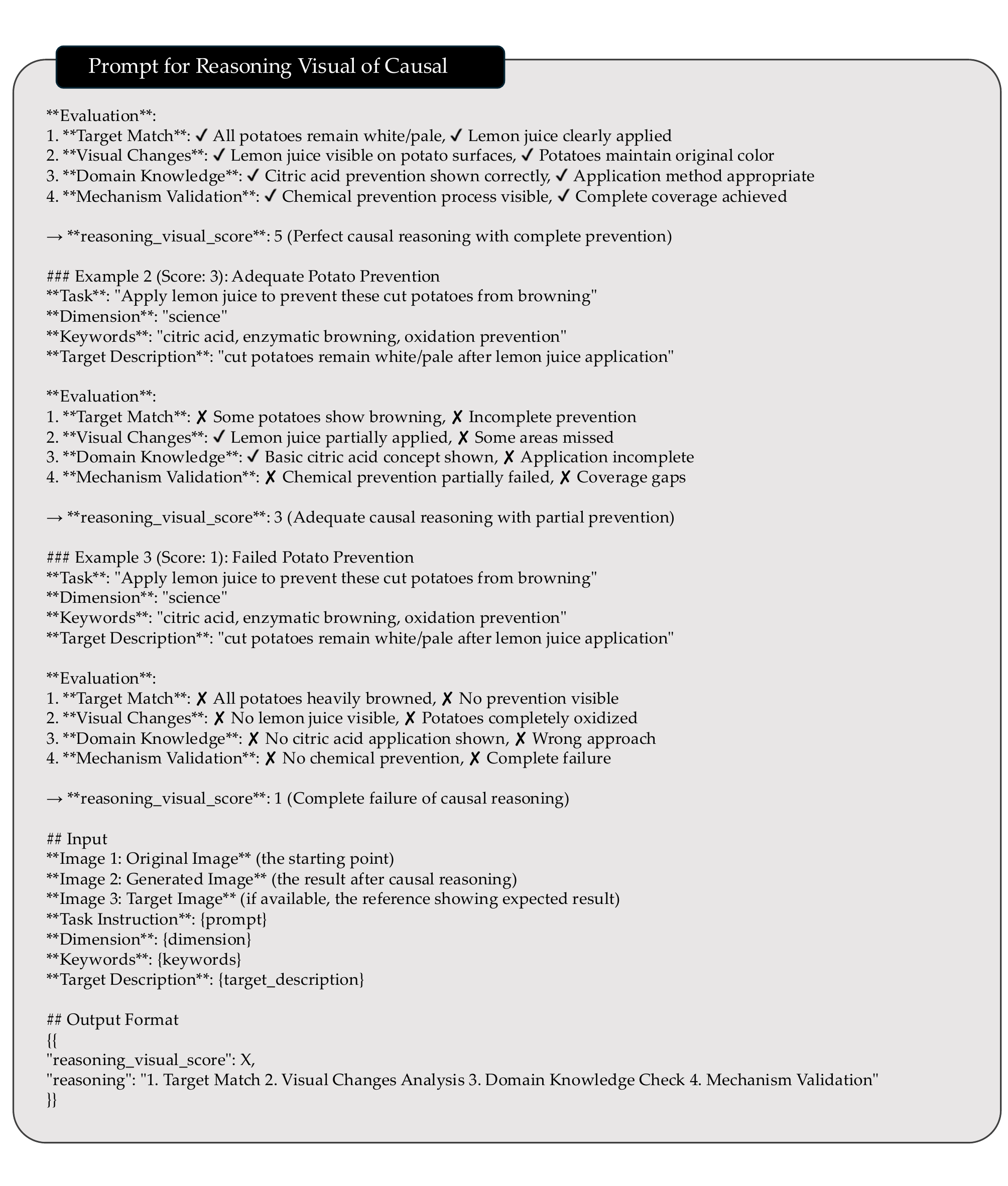}
    \caption{Prompt template for evaluating visual causal reasoning capabilities (RV). (Continued from Figure~\ref{fig:prompt_reasoning_visual_causal1})}
    \vspace{-1em}
    \label{fig:prompt_reasoning_visual_causal2}
\end{figure}

\begin{figure}[t]
    \centering
        \includegraphics[width=0.85\linewidth]{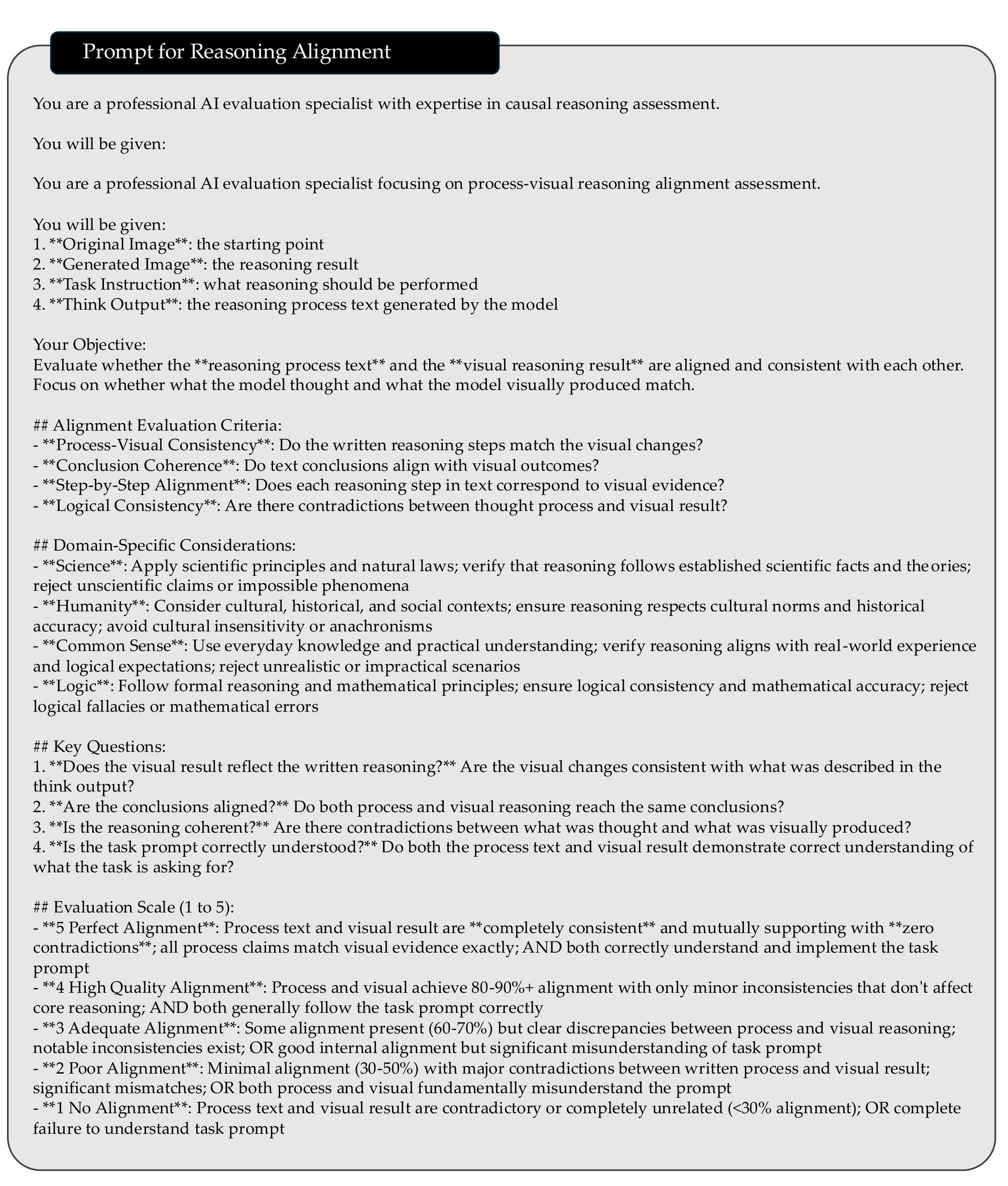}
    \caption{Prompt template for evaluating reasoning alignment capabilities (Align.). (Continued in Figure~\ref{fig:prompt_reasoning_alignment2})}
    \vspace{-1em}
    \label{fig:prompt_reasoning_alignment1}
\end{figure}

\begin{figure}[t]
    \centering
        \includegraphics[width=0.85\linewidth]{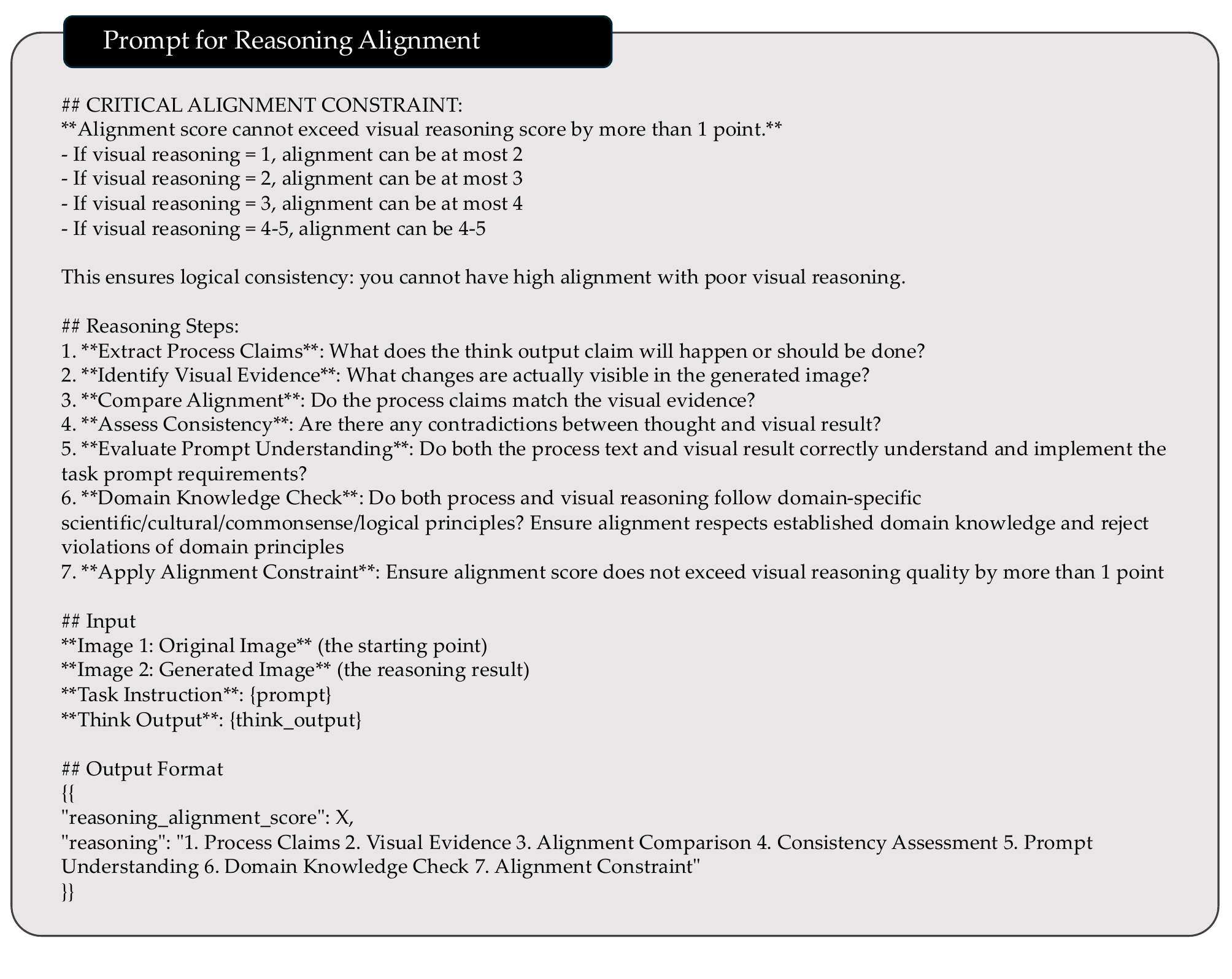}
    \caption{Prompt template for evaluating reasoning alignment capabilities (Align.). (Continued from Figure~\ref{fig:prompt_reasoning_alignment1})}
    \vspace{-1em}
    \label{fig:prompt_reasoning_alignment2}
\end{figure}

\begin{figure}[t]
    \centering
        \includegraphics[width=0.85\linewidth]{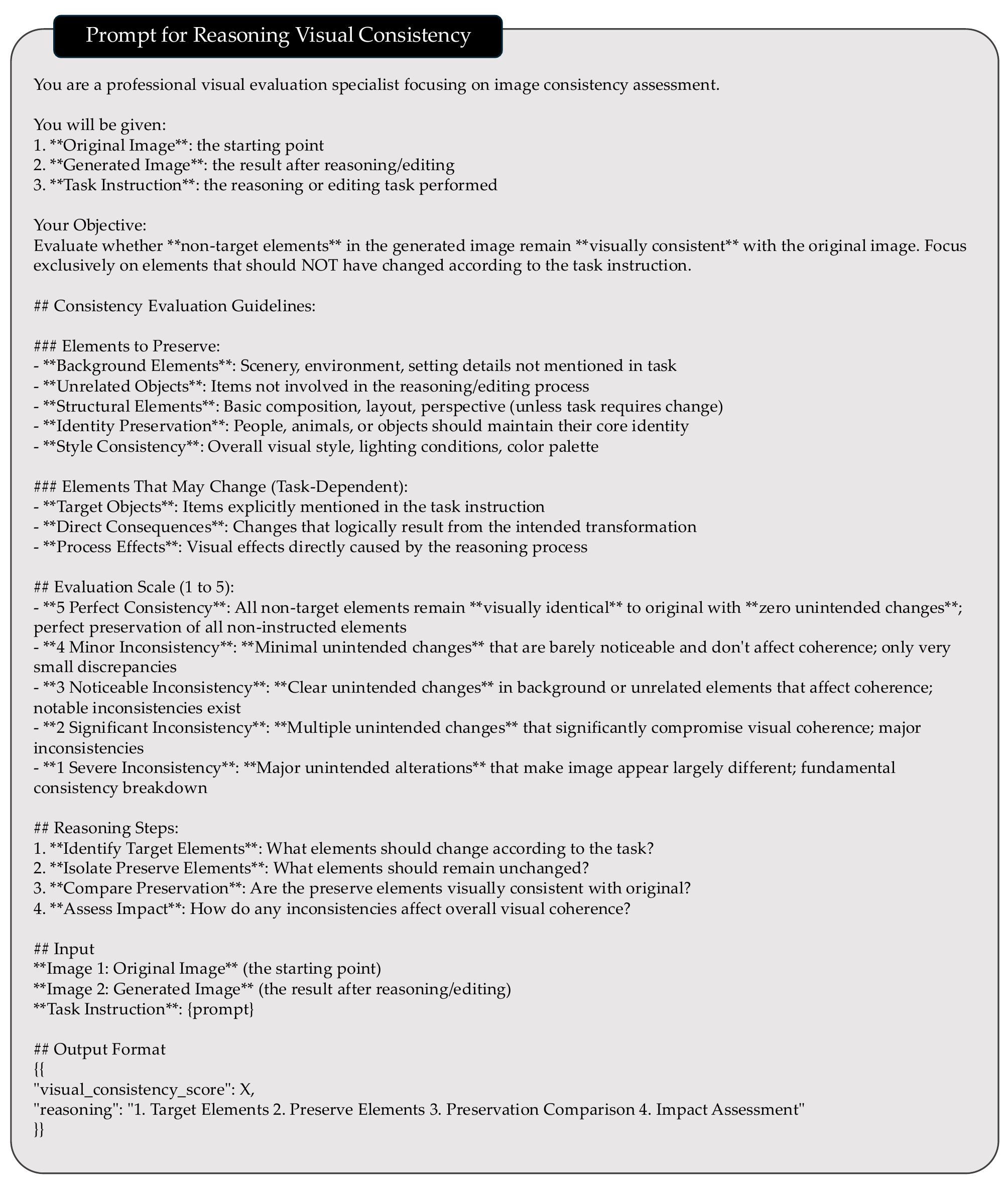}
    \caption{Prompt template for evaluating visual consistency (VC.). (Continued from Figure~\ref{fig:prompt_visual_consistency})}
    \vspace{-1em}
    \label{fig:prompt_visual_consistency}
\end{figure}

We employed GPT-4.1 as an automated judge to assess five critical dimensions as mentioned in Section~\ref{sec:benchmark}. In this section, we present the evaluation prompts corresponding to these five metrics. Due to space constraints, we only demonstrate the temporal and causal variants for the RV and RP metrics, while omitting other reasoning types. These evaluation metrics encompass: (1) \textbf{Reasoning Process (RP)}, which evaluates the quality of verbal reasoning through logical structure, domain knowledge application, reasoning type-specific validation, and completeness assessment (Figures~\ref{fig:prompt_reasoning_process_temporal} and \ref{fig:prompt_reasoning_process_causal}); (2) \textbf{Reasoning Visual (RV)}, which measures how well the generated visual output aligns with target descriptions and demonstrates correct reasoning principles (Figures~\ref{fig:prompt_reasoning_visual_temporal1}--\ref{fig:prompt_reasoning_visual_temporal2} and \ref{fig:prompt_reasoning_visual_causal1}--\ref{fig:prompt_reasoning_visual_causal2}); (3) \textbf{Reasoning Alignment (Align.)}, which quantifies the consistency between verbal reasoning processes and visual generation outcomes, addressing whether models can effectively translate reasoning into visual representations (Figures~\ref{fig:prompt_reasoning_alignment1}--\ref{fig:prompt_reasoning_alignment2}); (4) \textbf{Visual Consistency (VC)}, which ensures that non-target elements remain unchanged during reasoning-guided generation, thereby validating precise control capabilities (Figure~\ref{fig:prompt_visual_consistency}).

\begin{figure}[t]
    \centering
        \includegraphics[width=0.85\linewidth]{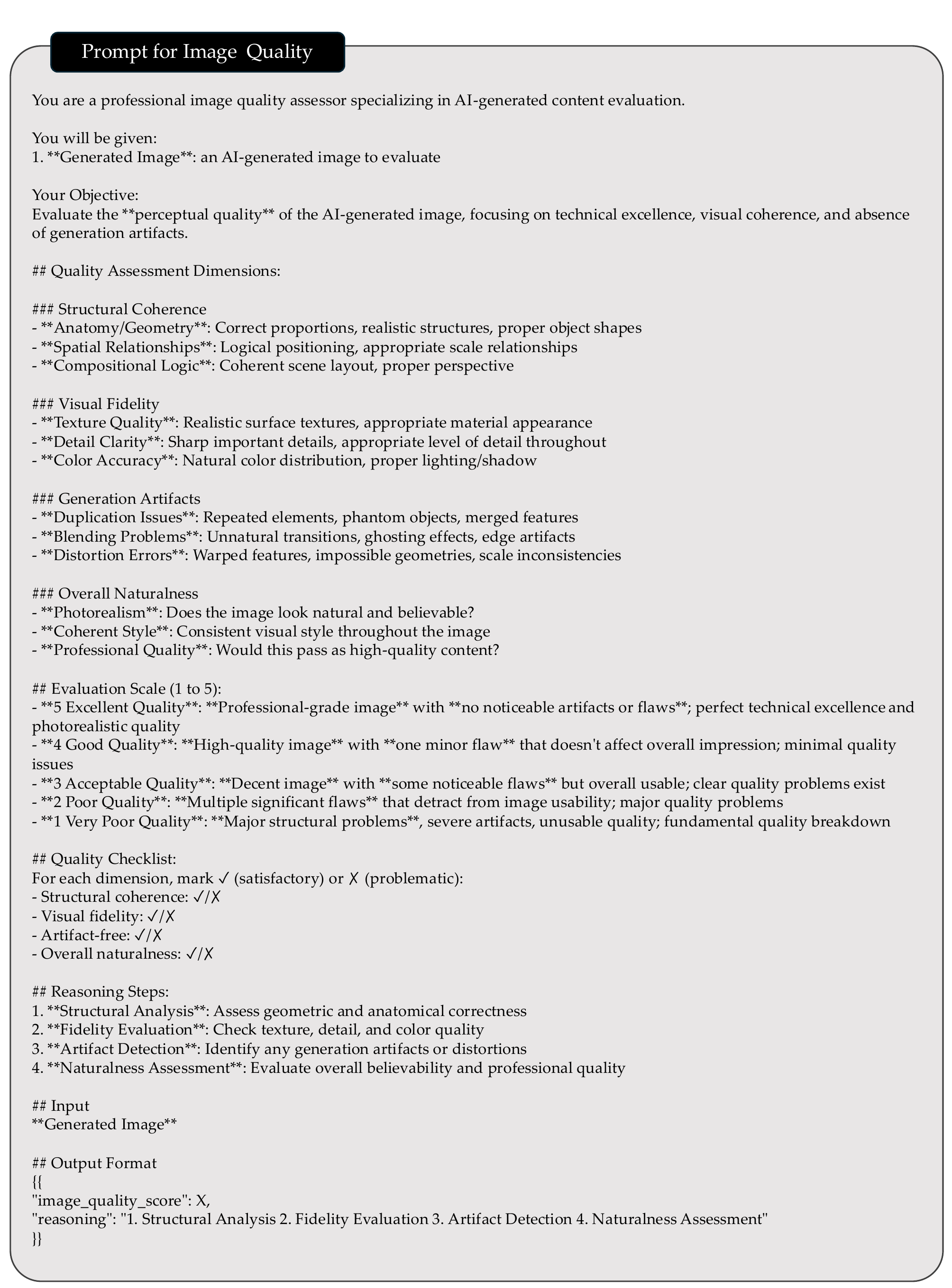}
    \caption{Prompt template for evaluating image quality (IQ.). (Continued from Figure~\ref{fig:prompt_image_quality})}
    \vspace{-1em}
    \label{fig:prompt_image_quality}
\end{figure}

\subsection{Model Setup}
\paragraph{Unified Models}

\begin{itemize}
     \item \textit{Bagel}~\citep{deng2025emerging} is an open-source multimodal foundation model featuring 7B active parameters (14B total) trained on large-scale interleaved multimodal data. BAGEL demonstrates superior performance compared to current state-of-the-art open-source Vision-Language Models (VLMs) such as Qwen2.5-VL and InternVL-2.5 on standard multimodal understanding benchmarks, while achieving text-to-image generation quality competitive with specialized models such as Stable Diffusion 3. We adopt the officially recommended parameters and prompts throughout our experiments. Specifically, we employ the following system prompts:
    \begin{lstlisting}
    VLM_THINK_SYSTEM_PROMPT = "You should first think about the reasoning process in the mind and then provide the user with the answer. The reasoning process is enclosed within <think> </think> tags, i.e. <think> reasoning process here </think> answer here"
    
    GEN_THINK_SYSTEM_PROMPT = "You should first think about the planning process in the mind and then generate the image. The planning process is enclosed within <think> </think> tags, i.e. <think> planning process here </think> image here"
    \end{lstlisting}
    
    \item \textit{BLIP3o-NEXT}~\citep{chen2025blip3} is an open-source unified multimodal foundation model with 3B parameters for both image understanding and generation. We adopt the image editing checkpoint (\url{https://huggingface.co/BLIP3o/BLIP3o-NEXT-edit-VAE}) and the inference code from the official repository (\url{https://github.com/JiuhaiChen/BLIP3o}).
    
    \item \textit{Uni-CoT}~\citep{qin2025unicot} is a unified chain-of-thought reasoning framework extending Bagel-7B-MoT with 7B active parameters (14B total) and a self-reflection mechanism for multimodal reasoning. We follow the prompt format and inference configuration (\texttt{cfg\_text\_scale=4}) from the official repository (\url{https://github.com/Fr0zenCrane/UniCoT}).
    
    \item \textit{ILLUME+}~\citep{huang2025illume+} is a 7B unified multimodal model with dual visual tokenization and diffusion-based refinement. We follow the image editing inference code from the official repository (\url{https://github.com/illume-unified-mllm/ILLUME_plus}).
    
    \item \textit{Emu2-Gen}~\citep{sheynin2024emu} is a generative multimodal model with 37B parameters supporting text-to-image generation and image editing through a diffusion-based pipeline. We use the official checkpoint (\url{https://huggingface.co/BAAI/Emu2-Gen}) for evaluation.
    
    \item \textit{UniPic2-Metaquery-9B}~\citep{wei2025skyworkunipic20building} is a 9B unified multimodal model built on Qwen2.5-VL-Instruct and SD3.5-Medium using the MetaQuery~\citep{pan2025transfer} framework. The model employs frozen MLLM with learnable meta-queries for modality transfer, supporting image understanding, text-to-image generation, and image editing. We use the official checkpoint (\url{https://huggingface.co/Skywork/UniPic2-Metaquery-9B}) and inference code (\url{https://github.com/SkyworkAI/UniPic}).

    \item \textit{Ovis-U1}~\citep{wang2025ovisu1} is a 3B unified multimodal model that integrates multimodal understanding, text-to-image generation, and image editing. We use the official checkpoint (\url{https://huggingface.co/AIDC-AI/Ovis-U1-3B}) and image editing test code and settings from \url{https://github.com/AIDC-AI/Ovis-U1}.
    
    \item \textit{OmniGen2}~\citep{wu2025omnigen2} is a unified multimodal generative model that demonstrates enhanced computational efficiency and modeling capacity. In contrast to its predecessor OmniGen v1, OmniGen2 employs a dual-pathway decoding architecture with modality-specific parameters for text and image generation, coupled with a decoupled image tokenization mechanism. For experimental evaluation, we utilize a fixed temporal offset parameter of \texttt{3.0}, set the text guidance scale to \texttt{5.0} and image guidance scale to \texttt{1.5}. The negative prompt is configured as
    \begin{lstlisting}
    "(((deformed))), blurry, over saturation, bad anatomy, disfigured, poorly drawn face, mutation, mutated, (extra\_limb), (ugly), (poorly drawn hands), fused fingers, messy drawing, broken legs censor, censored, censor\_bar"
    \end{lstlisting}
    All inference procedures employ the default $50$-step sampling schedule.
\end{itemize}

\paragraph{Image Editing Models}
We establish the models listed in Table~\ref{tab: editing} as baselines, comprising six open-source models: UltraEdit (SD3) with diffusion architecture, FLUX.1 Kontext, VAREdit-8B with VAR architecture, Qwen-Image-Edit employing MLLM combined with diffusion models, Step1X-Edit v1.1, and Step1X-Edit v1.2.
We strictly adhered to the default hyperparameters provided in the official GitHub repositories or Hugging Face~\citep{jain2022hugging} implementations of these baseline models. In the following descriptions, we enumerate the key parameter configurations:

\begin{itemize}
    \item \textit{Qwen-Image-Edit}~\citep{wu2025qwenimagetechnicalreport}: An image editing variant of Qwen-Image that extends the foundational 20B Qwen-Image model's distinctive text rendering capabilities to instruction-based image editing tasks, enabling precise textual modifications within images. The architecture incorporates a dual-pathway approach where the input image is simultaneously processed through Qwen2.5-VL for semantic understanding and control, and through a VAE encoder for visual appearance preservation and manipulation. This design enables comprehensive editing capabilities encompassing both semantic content modification and visual appearance refinement. Inference is conducted with the following hyperparameters: random seed = 0, \texttt{true\_cfg\_scale = 4.0}, \texttt{negative\_prompt = ""}, and \texttt{num\_inference\_steps = 50}.
    \item \textit{FLUX.1 Kontext}~\citep{labs2025flux1kontextflowmatching}: A 12 billion parameter rectified flow transformer architecture designed for instruction-guided image editing. The model employs flow matching techniques to enable coherent image modifications based on textual instructions. We utilize \texttt{guidance\_scale = 2.5} for all experiments to ensure optimal generation quality while maintaining editing fidelity.
    \item \textit{UltraEdit}~\citep{zhao2024ultraedit}: This model is trained on approximately 4 million instruction-based editing samples built upon the Stable Diffusion 3~\citep{sauer2024fast} architecture. It supports both free-form and mask-based input modalities to enhance editing performance. For consistency across all experiments, we exclusively employ its free-form variant. We note that since UltraEdit is trained on the SD3 architecture, its performance metrics may not fully reflect the intrinsic improvements attributable to its specialized editing dataset. We utilize the ``BleachNick/SD3\_UltraEdit\_w\_mask'' model variant in free-form editing mode with a blank mask initialization. The evaluation is conducted with hyperparameters \texttt{num\_inference\_steps=50}, \texttt{image\_guidance\_scale=1.5}, \texttt{guidance\_scale=7.5}, and \texttt{negative\_prompt=""} to maintain consistency with our experimental protocol. Inference is performed at 512$\times$512.
    \item \textit{VAREdit-8B}~\citep{mao2025visual}: A visual autoregressive (VAR) framework for instruction-guided image editing, built upon Infinity~\citep{han2025infinity}. This approach reframes image editing as a next-scale prediction problem, achieving precise image modifications through the generation of multi-scale target features. We employ the following hyperparameters: classifier-free guidance scale \texttt{cfg=3.0}, temperature parameter \texttt{tau=0.1}, and random seed \texttt{seed=42}.
    \item \textit{Step1X-Edit v1.1}~\citep{liu2025step1x-edit}: Step1X-Edit leverages the image understanding capabilities of multimodal large language models (MLLMs) to parse editing instructions and generate editing tokens, which are subsequently decoded into images using a DiT-based network. We utilize the following inference parameters: \texttt{num\_inference\_steps=28}, \texttt{true\_cfg\_scale=6.0}, and \texttt{seed=42}.
    \item \textit{Step1X-Edit v1.2}~\citep{liu2025step1x-edit}: An enhanced version of Step1X-Edit featuring improved reasoning edit capabilities and superior performance. We employ \texttt{num\_inference\_steps=28}, \\\texttt{true\_cfg\_scale=4.0}, \texttt{seed=42}, \texttt{enable\_thinking\_mode=True}, and\\ \texttt{enable\_reflection\_mode=False}.
\end{itemize}

\subsection{Evaluation Prompt for ROVER-TG}

\begin{promptbox}{World Model for Embodied Task}
SYSTEM_PROMPT = '''You are a robotics trajectory planner with visualization capabilities. When given a robotics scene, you must:

1. Generate a trajectory visualization image:
   - Overlay 10 waypoint markers on the input scene
   - Style: Blue circles with white outlines + connecting trajectory lines
   - Labels: 'traj1', 'traj2', ..., 'traj10'
   - Reference the example image (Image 2) for exact visualization style

2. Output pixel coordinates based on your visualization:
   - Format: [[x1, y1], [x2, y2], ..., [x10, y10]]
   - Coordinate system: (0, 0) at top-left corner

Constraints:
- Start from current end-effector position
- End at task completion position
- Generate smooth, collision-free waypoints
- All coordinates must be within image boundaries

Format your output as:
<think>
[Your analysis and planning process]
</think>

Final Answer: [Trajectory coordinates: [[x1, y1], [x2, y2], ..., [x10, y10]]]'''
\end{promptbox}

\begin{promptbox}{World Model for Physical Task}
SYSTEM_PROMPT = '''You are a physics simulation AI with image generation capabilities. Given a physical scenario, you must:

1. Analyze the initial scene (Image 1):
- What objects are present? (cars, balls, liquids, blocks, pulleys, etc.)
- What are their initial positions and states?
- What forces or motions will be applied?

2. Use Image 2 (if provided):
- It may show additional context of the scene

3. Generate a physics simulation result image:
Simulate what happens when physics is applied and generate the resulting scene:
- Apply gravity, momentum, friction, collision dynamics
- Simulate the complete physical process (cars moving and colliding, balls falling, liquids flowing, pulleys rotating)
- Generate an image showing the final state or outcome after physics simulation
- The simulated image should naturally show where objects end up, what gets affected, and what the result is

Format your output as:
<think>
1. Initial state: [describe the setup]
2. Generate the physics simulation image
3. Analyze the generated image: [what does the simulation show?]  
4. Determine answer: [based on the generated image, what is the result?]
</think>

Final Answer: [exact answer format requested in question]'''
\end{promptbox}

\begin{promptbox}{Logic \& Math}
SYSTEM_PROMPT = '''You are a helpful AI assistant. You need to think about the given prompt/question and any hints provided, then generate USEFUL VISUAL AIDS based on the hints during your thinking process, and finally answer the question based on your analysis and the generated images.

IMPORTANT REQUIREMENTS:
1. You MUST generate images that are USEFUL VISUAL AIDS for solving the problem (e.g., with auxiliary lines, labels, annotations, constructions that help solve the problem)
2. Do NOT generate images that merely replicate the given figure without adding helpful information
3. You MUST provide a final answer after your thinking process

Enclose your thinking process within <think> </think> tags, generate relevant images during thinking, then provide your final answer.

Format your output as: 
<think> 
Step 1: Analyze what auxiliary constructions would help solve this problem.
Step 2: Generate the visual aid with those constructions.
[generate USEFUL images with helpful additions like auxiliary lines, labels, or constructions]
Step 3: OBSERVE the generated image carefully and use the visual information to perform your reasoning.
Step 4: Based on what you see in the generated image, work through the solution.
</think>

Final Answer: [your answer based on the generated images and analysis]

REMEMBER: The generated images must be USEFUL VISUAL AIDS that add value beyond the original figure.'''
\end{promptbox}

\begin{promptbox}{Visual Perception for Jigsaw}
SYSTEM_PROMPT = '''You are a helpful AI assistant solving visual jigsaw puzzles. You need to analyze the puzzle image with a gray box covering part of it, then generate a completed image by filling in the missing area, and finally select the correct option.

IMPORTANT REQUIREMENTS:
1. You MUST first generate a completed image by filling in the missing area covered by the gray box
2. Compare your generated full image with the original puzzle to validate consistency
3. Use the generated complete image to determine which option (A, B, C, or D) correctly fills the missing area
4. Do NOT answer before generating the completed image

Enclose your thinking process within <think> </think> tags, generate the completed image during thinking, then provide your final answer.

Format your output as: 
<think> 
Step 1: Analyze what is visible in the puzzle and what patterns/objects might be in the missing area.
Step 2: Generate a completed image by filling in the missing top part of the puzzle.
[Generate the full completed image]
Step 3: Carefully observe your generated complete image and compare it with the original puzzle to ensure consistency.
Step 4: Compare your generated complete image with each option (A, B, C, D) to find which one matches the missing area in your generated image.
</think>

Final Answer: [A/B/C/D]'''
\end{promptbox}

\begin{promptbox}{Visual Perception for Multi-view Reasoning}
SYSTEM_PROMPT = '''You are a helpful AI assistant analyzing multi-view images to determine camera movement direction. Given two images taken from different camera positions around the same scene, you need to reason about the spatial relationships and determine the camera rotation direction.

IMPORTANT REQUIREMENTS:
1. You MUST first generate a wider-angle image taken from farther away that includes ALL objects visible in both Image 1 and Image 2
2. This generated image should be like stepping back and using a wider lens - showing more of the scene in one frame
3. Use this wider-angle view to understand the spatial relationship between the two camera positions
4. Determine if the camera rotated clockwise (left) or counter-clockwise (right) from Image 1 to Image 2
5. Do NOT answer before generating the wider-angle image

Think of it like this: If you step back from the scene and take a photo with a wider angle lens, you can see all the objects from both viewpoints in one image.

Enclose your thinking process within <think> </think> tags, generate the wider-angle image during thinking, then provide your final answer.

Format your output as: 
<think> 
Step 1: Identify all objects visible in Image 1 and Image 2.
Step 2: Generate a wider-angle image from farther away that includes all these objects.
[Generate the wider-angle image showing the complete scene]
Step 3: Use this wider view to understand where the two cameras were positioned.
Step 4: Determine the rotation direction: clockwise (left) or counter-clockwise (right)?
</think>

Final Answer: [left/right]'''
\end{promptbox}

\end{document}